\documentclass[sigconf]{acmart}
\AtBeginDocument{%
  \providecommand\BibTeX{{%
    \normalfont B\kern-0.5em{\scshape i\kern-0.25em b}\kern-0.8em\TeX}}}

\copyrightyear{2023}
\acmYear{2023}
\setcopyright{rightsretained}
\acmConference[KDD '23]{Proceedings of the 29th ACM SIGKDD Conference on Knowledge Discovery and Data Mining}{August 6--10, 2023}{Long Beach, CA, USA}
\acmBooktitle{Proceedings of the 29th ACM SIGKDD Conference on Knowledge Discovery and Data Mining (KDD '23), August 6--10, 2023, Long Beach, CA, USA} 
\acmDOI{10.1145/3580305.3599833}
\acmISBN{979-8-4007-0103-0/23/08}

\usepackage{mathtools}
\usepackage{multirow}
\usepackage{soul}
\usepackage{wrapfig}
\usepackage{caption}
\usepackage{subcaption}
\usepackage{hyperref}
\usepackage{xspace}
\usepackage{colortbl}
\usepackage{multicol}
\usepackage{footnote}
\usepackage{tablefootnote}
\usepackage{threeparttable}
\usepackage{enumitem}
\usepackage{balance}

\definecolor{Gray}{gray}{0.85}

\DeclareMathOperator*{\argmin}{arg\,min}

\DeclareMathOperator*{\softmax}{softmax}

\newcommand{\amzcore}{\texttt{Amazon-PQ}\xspace}
\newcommand{\amzreviews}{\texttt{Product-Reviews}\xspace}
\newcommand{\searchctr}{\texttt{Search-CTR}\xspace}
\newcommand{\esci}{\texttt{ESCvsI}\xspace}
\newcommand{\qpt}{\texttt{Query2PT}\xspace}
\newcommand{\copurchase}{\texttt{CoPurchase}\xspace}
\newcommand{\acat}{\texttt{Product2PT}\xspace}

\newcommand{\galm}{\textsc{GaLM}\xspace}
\newcommand{\galmco}{$\mathrm{\textsc{GaLM}^{co}}$\xspace}

\newcommand{\lmgnn}{\textsc{LM+GNN}\xspace}

\newcommand{\hf}{$\mathrm{\textsc{LM}_{\textsc{(Bert-base)}}}$\xspace}
\newcommand{\hfft}{$\mathrm{\textsc{LM}_{\textsc{(Bert-base)}}^*}$\xspace}
\newcommand{\bertmlm}{$\mathrm{\textsc{LM}_{\textsc{(Bert-mlm)}}}$\xspace}
\newcommand{\bertwg}{\textsc{GaLM}\xspace}
\newcommand{\bertwgft}{$\mathrm{\textsc{GaLM}^*}$\xspace}
\newcommand{\bertcorgcn}{$\mathrm{\textsc{GaLM}^{co_{(rgcn)}}}$\xspace}
\newcommand{\bertcorgat}{$\mathrm{\textsc{GaLM}^{co_{(rgat)}}}$\xspace}

\newcommand{\galmrgat}{$\mathrm{\textsc{GaLM}_{rgat}}$\xspace}
\newcommand{\galmrgnnft}{$\mathrm{\textsc{GaLM}^*_{rgnn}}$\xspace}
\newcommand{\galmrgcnft}{$\mathrm{\textsc{GaLM}^*_{rgcn}}$\xspace}
\newcommand{\galmrgatft}{$\mathrm{\textsc{GaLM}^*_{rgat}}$\xspace}

\newcommand{\galmrgnnftplus}{$\mathrm{\textsc{GaLM}^*_{rgnn}}$+\xspace}
\newcommand{\galmrgcnftplus}{$\mathrm{\textsc{GaLM}^*_{rgcn}}$+\xspace}
\newcommand{\galmrgatftplus}{$\mathrm{\textsc{GaLM}^*_{rgat}}$+\xspace}
\newcommand{\galmrgatftplusft}{$\mathrm{\textsc{GaLM}^{**}_{rgat}}$+\xspace}

\newcommand{\rgat}{$\mathrm{\textsc{Gnn}_{(rgat)}}$\xspace}

\begin{document}

\title{Graph-Aware Language Model Pre-Training on a Large Graph Corpus Can Help Multiple Graph Applications}

\author{Han Xie}
\authornote{This work was done during Han Xie’s internship at Amazon, USA.}
\affiliation{%
  \institution{Emory University}
  \city{Atlanta}
  \state{GA}
  \country{USA}
}
\email{han.xie@emory.edu}

\author{Da Zheng}
\affiliation{%
  \institution{Amazon AWS AI}
  \city{Santa Clara}
  \state{CA}
  \country{USA}
}
\email{dzzhen@amazon.com}

\author{Jun Ma}
\authornote{Work done while at Amazon.}
\affiliation{%
  \institution{Walgreens AI Lab}
  \city{Bellevue}
  \state{WA}
  \country{USA}
}
\email{jun.ma@walgreens.com}

\author{Houyu Zhang}
\affiliation{%
  \institution{Amazon Search AI}
  \city{Seattle}
  \state{WA}
  \country{USA}
}
\email{zhanhouy@amazon.com}

\author{Vassilis N. Ioannidis}
\affiliation{%
  \institution{Amazon Search AI}
  \city{Santa Clara}
  \state{CA}
  \country{USA}
}
\email{ivasilei@amazon.com}

\author{Xiang Song}
\affiliation{%
 \institution{Amazon AWS AI}
  \city{Santa Clara}
  \state{CA}
   \country{USA}
 }
\email{xiangsx@amazon.com}

\author{Qing Ping}
\affiliation{%
 \institution{Amazon Search AI}
 \city{Palo Alto}
  \state{CA}
   \country{USA}
 }
\email{pingqing@amazon.com}

\author{Sheng Wang}
\affiliation{%
  \institution{Amazon Scholar}
  \city{Seattle}
  \state{WA}
  \country{USA}}
\email{swanguw@amazon.com}

\author{Carl Yang}
\affiliation{%
  \institution{Emory University}
  \city{Atlanta}
  \state{GA}
  \country{USA}}
\email{j.carlyang@emory.edu}

\author{Yi Xu}
\affiliation{%
  \institution{Amazon Search AI}
  \city{Seattle}
  \state{WA}
  \country{USA}
  }
\email{yxaamzn@amazon.com}

\author{Belinda Zeng}
\affiliation{%
  \institution{Amazon Search AI}
  \city{Seattle}
  \state{WA}
  \country{USA}
  }
\email{zengb@amazon.com}

\author{Trishul Chilimbi}
\affiliation{%
  \institution{Amazon Search AI}
  \city{Seattle}
  \state{WA}
  \country{USA}
  }
\email{trishulc@amazon.com}

\renewcommand{\shortauthors}{Xie, et al.}

\begin{abstract}
Model pre-training on large text corpora has been demonstrated effective for various downstream applications in the NLP domain. In the graph mining domain, a similar analogy can be drawn for pre-training graph models on large graphs in the hope of benefiting downstream graph applications, which has also been explored by several recent studies. However, no existing study has ever investigated the pre-training of text plus graph models on large heterogeneous graphs with abundant textual information (a.k.a. large graph corpora) and then fine-tuning the model on different related downstream applications with different graph schemas. To address this problem, we propose a framework of graph-aware language model pre-training (\galm) on a large graph corpus, which incorporates large language models and graph neural networks, and a variety of fine-tuning methods on downstream applications. We conduct extensive experiments on Amazon's real internal datasets and large public datasets. Comprehensive empirical results and in-depth analysis demonstrate the effectiveness of our proposed methods along with lessons learned. 

\end{abstract}





\maketitle

\section{Introduction}
\label{intro}

The standard process of pre-training a large language model (LM) on abundant text data and fine-tuning the pre-trained model on different application data has achieved significant success and brought revolutionary advancement to the domain of natural language processing. With the massive corpus and powerful computation resources for pre-training, the pre-trained LMs based on powerful transformer architectures emerge and derive various families \cite{min2021recent}, including auto-regressive LMs like GPT \cite{radford2018improving} and GPT-2/3/4 \cite{radford2019language, brown2020language, openai2023gpt4}, masked LMs like BERT \cite{devlin2018bert}, RoBERTa \cite{liu2019roberta}, and XLNet \cite{yang2019xlnet}, and encoder-decoder LMs like BART \cite{lewis2020bart} and T5 \cite{raffel2020exploring}. These pre-trained LMs can be directly fine-tuned on users' data for downstream applications to achieve higher utility and/or efficiency. However, for enterprises that usually preserve their own in-domain data and target diverse applications, existing LMs that are pre-trained on text in the general domain can be less useful. Thus, for an enterprise with sufficient in-domain data, it is practical and desirable to pre-train its own large LMs, such as by further training existing general LMs. 

In addition to pure text data, graph-structured data have been used increasingly in the industry to model the complex real-world relations between entities. For example, in a recommender system, users and items can be represented as two types of nodes in a graph, and user behaviors (e.g., purchases, likes) can be represented as various types of edges connecting nodes, forming a large heterogeneous network \cite{yang2020heterogeneous}. Thus, realistic applications such as predicting whether a customer would buy an item can be modeled as graph downstream tasks, e.g., link prediction. With the advancement of graph representation learning techniques, graph neural network (GNN) and its variants become popular and aid in the learning of graph-based applications, such as GCN \cite{kipfsemi}, GraphSAGE \cite{hamilton2017inductive}, GIN \cite{xu2018how}, and GAT \cite{busbridge2019relational} for homogeneous graphs with single node and edge types, and HGT \cite{hu2020heterogeneous}, HAN \cite{wang2019heterogeneous}, RGCN \cite{schlichtkrull2018modeling}, and RGAT \cite{busbridge2019relational} for heterogeneous graphs with multiple node and edge types.

In real scenarios, it is a common practice for an enterprise or an institute to own a domain-specific \textit{graph corpus}, that is, a massive graph with abundant text information as node features. In the meantime, different departments collect specific graph corpora due to the diverse commercial or research needs. For example, in Amazon, users' interactions with different entities (e.g., products, queries, ads) can be collected through the e-commerce engine and used to construct a large graph corpus, which is a heterogeneous graph with rich text information (e.g., users' reviews, products' descriptions). Meanwhile, different departments that develop diverse applications with different commercial objectives (e.g., predicting the clickthrough rate for advertising, semantic matching between queries and products), collect data based on their unique data access and application need and construct the corresponding application graph corpus that can include both overlapping and distinct parts from the large graph corpus. Considering utility and efficiency benefits, it meets the real need of enterprises to leverage the large graph corpus to facilitate various applications. Another example can be found in the biomedical domain, where large-scale protein-protein interaction networks might be massively measured on model organisms such as mouse and yeast. However, experimentally deriving large-scale networks on other species could be technically challenging and expensive. As a result, there is a pressing need to pre-train models on large well-studied species and then transfer them to small-scale networks on other species. Motivated by the successes of LM pre-training on large text corpora, the emerging data resource of large graph corpora, and the need of facilitating multiple downstream graph-based applications, there is an urge for methodology design that can effectively utilize the large graph corpora to promote the performance of various downstream applications.

In this work, we focus on a setting where there exists a large graph corpus with multiple types of entities and relations along with their rich textual information, and various downstream graph-based applications whose relation types can be both overlapping and distinct from the large graph corpus. To enable this, we consider the technical challenges of how to learn a powerful model on a large graph corpus, and how to transfer it to downstream application graphs. 
Regarding learning on a graph corpus, previous studies leverage techniques that combine LMs and GNNs, which usually encode the text information through LMs and feed the outputs of LMs to GNNs as node features, such as TextGNN \cite{zhu2021textgnn}, AdsGNN \cite{li2021adsgnn}, and GIANT \cite{chiennode}. However, these methods are designed for specific scenarios or focus only on downstream applications.
As for transferring the knowledge of large graphs, most previous studies focus on pre-training and fine-tuning over the same graphs for different tasks, e.g., designing self-supervised training methods without labeled data \cite{jiang2021contrastive, hwang2020self}, or transferring to the applications that have the same graph schema (node and edge types) as the pre-training graph, e.g., using external knowledge graphs \cite{rosset2020knowledge, yasunagadeep}. Thus they are inapplicable to the setting of transferring information across graphs with diverse graph schemas. Moreover, these works concentrate on the transfer learning of GNNs which neglect the impact of text-based node features on graph topology and vice versa. To the best of our knowledge, there exists no prior attempt at model pre-training on a large graph corpus and applying the pre-trained model to multiple applications, where the tasks of applications can vary and the graph schemas of applications can differ from the large pre-training graph.

\begin{figure}
     \centering
     \begin{subfigure}[b]{0.47\columnwidth}
         \centering
         \includegraphics[width=\columnwidth]{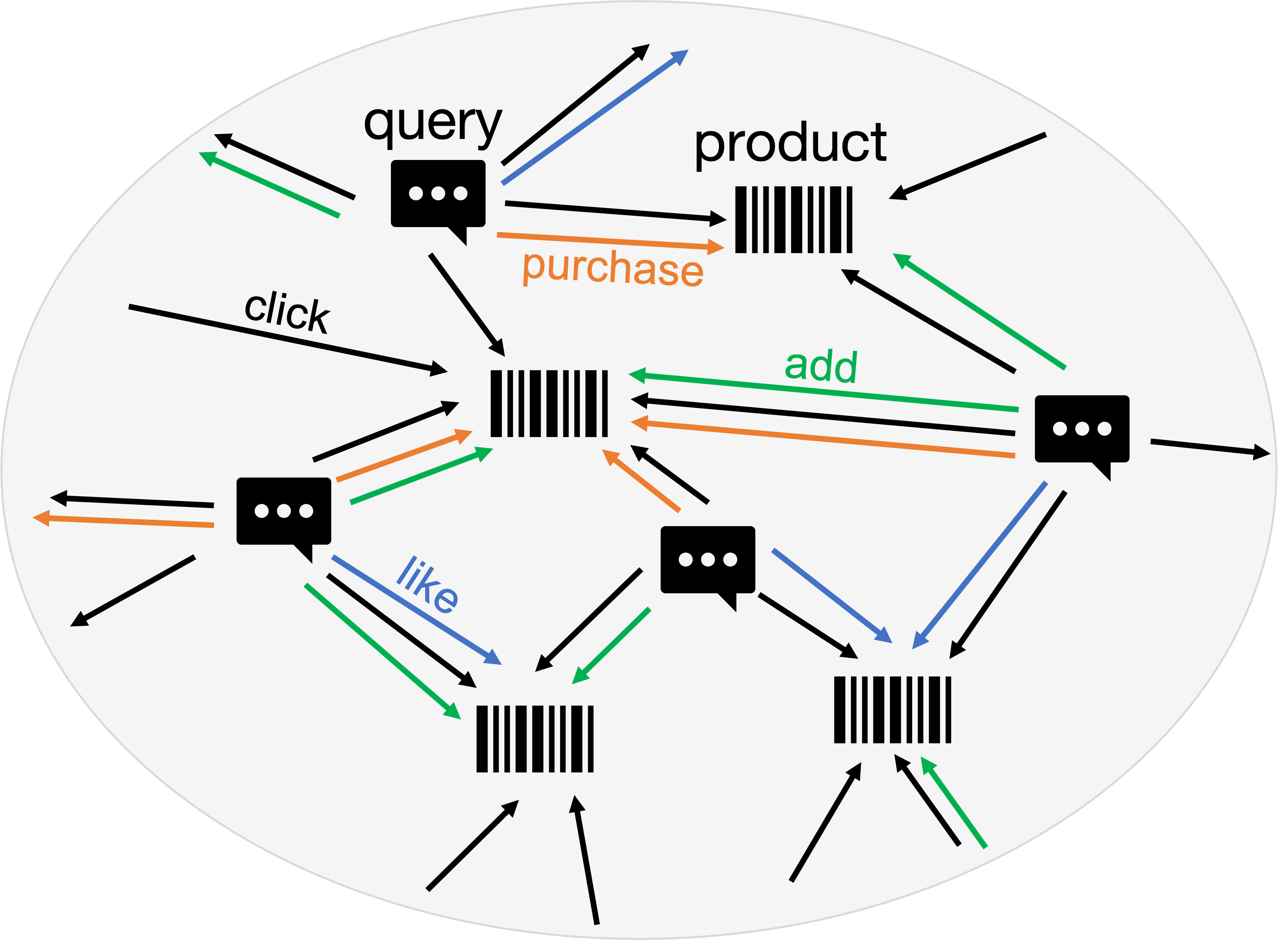}
         \caption{A large graph corpus}
         \label{fig:coregraph}
     \end{subfigure} \hfill
     \begin{subfigure}[b]{0.47\columnwidth}
         \centering
         \includegraphics[width=\columnwidth]{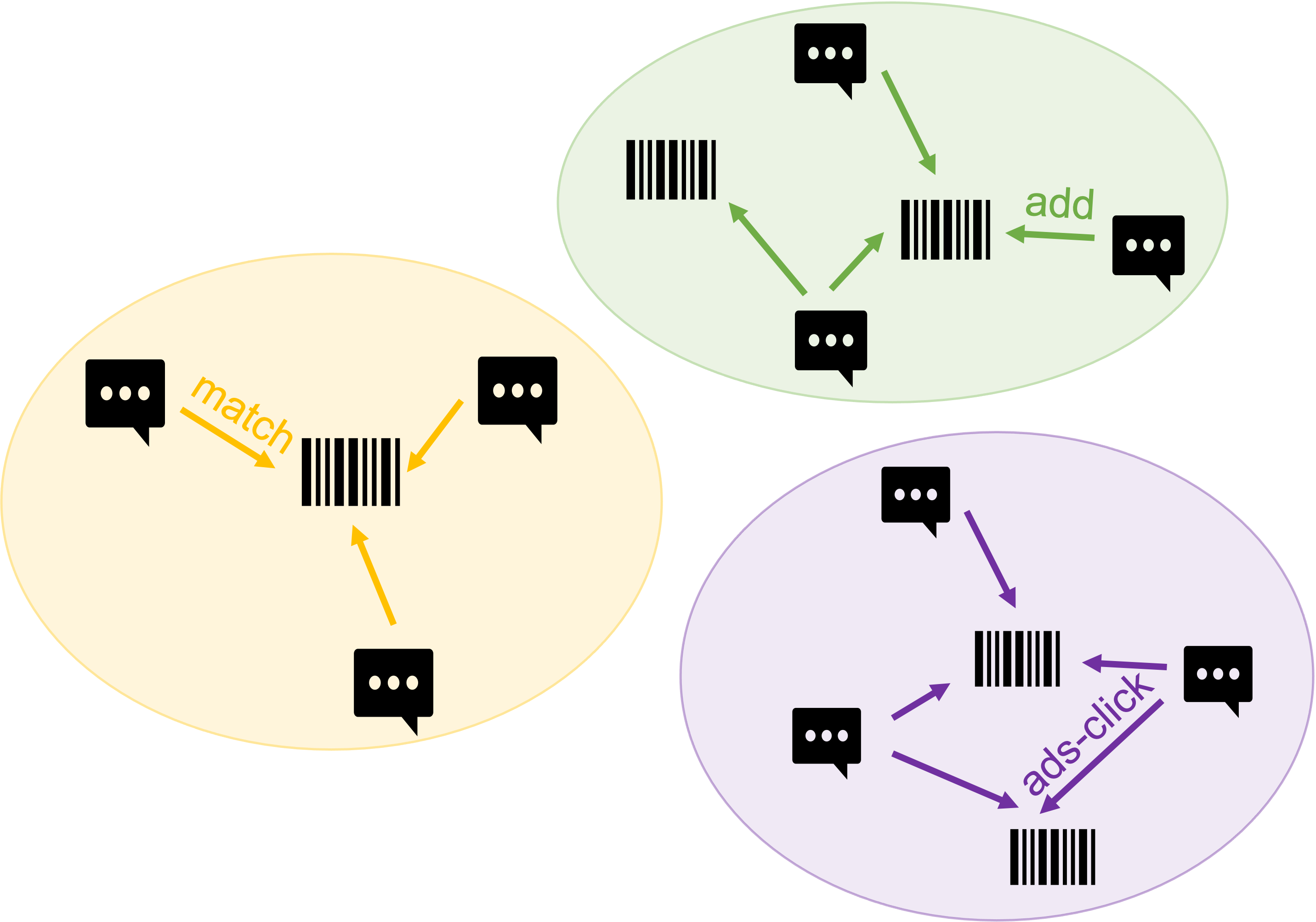}
         \caption{Three application graphs}
         \label{fig:appgraph}
     \end{subfigure}
    \caption{Toy examples of a large graph corpus and application graphs. \textmd{\small{The large graph covers most of the nodes in applications, while the edge types of applications can be distinct from the large graph.}}}
\label{fig:graphs}
\end{figure}

We approach the key problem from two perspectives, pre-training a powerful model on a large graph corpus to capture the information that can maximize its utility towards a variety of applications, and fine-tuning the pre-trained model on various applications to further enhance its performance. Our main contributions include:
\begin{itemize}[leftmargin=*]
    \item We propose a framework of \textbf{G}raph-\textbf{A}ware \textbf{LM} pre-training on a large graph corpus, a.k.a. \textbf{\galm}, which can encode knowledge in the large graph corpus into LMs under the consideration of entity relations in graphs.
    \item We propose various methods for fine-tuning the pre-trained \galm on applications w.r.t.~different modules of \galm.
    \item We simulate a large graph corpus and two applications from a large public dataset; the comprehensive experiments on Amazon internal datasets and the public datasets demonstrate the effectiveness of our proposed pre-training/fine-tuning strategies.
    \item We provide insights into the empirical results, as well as additional analysis and discussion over the lessons learned.
\end{itemize}

In the rest of this paper, we discuss related works in Section \ref{sec:relworks}. Section \ref{sec:preliminaries} introduces preliminaries including the problem formulation, the prevalent yet new concept of large graph corpus, and multiple related real applications that are deployed in Amazon. We then introduce the backbone model \lmgnn of our framework in Section \ref{sec:lmgnn}. Sections \ref{sec:pretraining} and \ref{sec:finetuning} illustrate our proposed pre-training framework and fine-tuning strategies, respectively, and include empirical results and insights. Lastly, we provide an overall comparison and additional analysis in Section \ref{sec:exprs}.

\section{Related Works}
\label{sec:relworks}

\subsection{Language Modeling with Graphs}

Learning on graphs has attracted significant attention recently due to its capability of leveraging both node features and topological information to better model real-world applications. For language modeling, introducing the topological information that either models the real interactions (e.g., human behaviors) or semantic linkages (e.g., knowledge graphs, context) helps to overcome the inadequacy of utilizing solely pure semantic information. For example, \cite{shen2020exploiting, wang2021kepler, yu2022jaket, ke2021jointgt, yasunagadeep} leverage KGs as additional training signals to guide the learning of LMs, \cite{zhu2021textgnn, li2021adsgnn} use behavior-graph augmented language modeling that is applied for sponsored search, and \cite{yao2019graph, menggnn} construct graphs based on the raw text and represent the tokens or words as nodes.

Graph neural networks (GNNs) are widely-used techniques in graph representation learning, and some previous works of learning LMs with graphs employ GNNs to aggregate information that is propagated along the graph topology. However, how LMs and GNNs are combined can vary. For example, \cite{menggnn} encodes tokens using a vanilla neural LM and introduces the global context by constructing a heterogeneous graph with tokens as nodes and connecting tokens to their retrieved neighbors, which essentially stacks the GNNs on top of the vanilla LM without fine-tuning the LM. The LM performs the role of a static text encoder in this way of combining LMs and GNNs. \cite{zhu2021textgnn, li2021adsgnn} both focus on sponsored search, of which \cite{li2021adsgnn} first fine-tunes the pre-trained LMs by behavior-graph related tasks but still uses the fine-tuned LMs as static text encoders to generate fixed node features as the input into GNNs, while \cite{zhu2021textgnn} leverages the behavior-graph as a complementary information and co-trains the parameters of LMs and GNNs. \cite{ioannidis2022efficient} proposes a multi-step fine-tuning framework for LMs that can jointly train LMs and GNNs effectively and efficiently. 

Different from the prior works, this work is generic and not application-specific, and not restricted to specific graph schema such as KGs. More importantly, it focuses on LM pre-training on a large graph corpus with graph information included and fine-tuning the pre-trained model on multiple applications where the edge schema can be distinct from the pre-training large graph.

\subsection{Pre-training on Graphs}
Many previous studies of pre-training on graphs concentrate on encoding the graph-level information into GNNs given enormous small graphs, especially in the molecule domain (e.g., \cite{Hu2020Strategies, rong2020self, sun2022does}). Apart from the graph-level pre-training, there emerge works studying pre-training on a large graph, which aim to address the problem of insufficient labels by self-supervised learning on the same graph (e.g., \cite{jiang2021contrastive, hwang2020self, hu2020gpt}), or to tailor the gap of optimization objectives and training data between self-supervised pre-training tasks and downstream tasks (e.g., {\cite{han2021adaptive, lu2021learning}}), or to pre-train on a large graph to capture the transferable information for downstream application graphs (e.g., \cite{zhu2021transfer, lu2021learning, qiu2020gcc, fang2020pre, jiang2021pre}). Our setting is more proximate to the last category.

Among these works concerning transfer learning across graphs, \cite{zhu2021transfer} provides theoretical analysis for the transferability of GNNs and proposes a GNN transfer learning framework with ego-graph information maximization, \cite{lu2021learning} proposes a self-supervised pre-training strategy that learns fine-tuning during pre-training with a dual adaption mechanism at both node and graph levels, \cite{qiu2020gcc} learns a universal and generic GNN model that can be applied to data from diverse domains and different tasks using contrastive learning (subgraph instance discrimination). However, all these works focus on homogeneous graphs. \cite{jiang2021pre} first studies the pre-training of GNN on a large-scale heterogeneous graph with both node- and schema-level contrastive learning tasks that can be applied to in-domain new datasets. \cite{fang2020pre} proposes a generic pre-training framework for heterogeneous graphs by transforming the neighbors into sequences and adopting deep bi-directional transformers to encode them, under the supervision of masked node modeling and adjacent node prediction. However, these works concentrate on transferring across graphs with the same graph schema, while in our setting the application graphs can preserve new edge types distinct from the pre-training graph, which can enrich the set of applicable downstream tasks (e.g., predicting the new edge type) and introduce specific features to applications by involving new edge types. Moreover, most of the previous studies on pre-training with graphs omit the relationship between the raw text and graph topology and use node features that are generated in a graph-agnostic manner. In this work, we focus on leveraging graphs and GNNs to facilitate the information capturing of LMs on a large graph corpus.

\section{Preliminaries}
\label{sec:preliminaries}

\subsection{Problem Formulation}
Given a large-scale heterogeneous graph with text information (the large graph corpus) $G_\mathrm{c}$, and multiple downstream application heterogeneous graphs $\mathcal{G} = \{g_i\}_{i=1}^{n}$, the goal is to leverage $G_\mathrm{c}$ to improve tasks on $g_i$.

We denote the large graph corpus as $G_\mathrm{c} = (V_\mathrm{c}, E_\mathrm{c}, A_\mathrm{c}, R_\mathrm{c}, \phi_\mathrm{c}, \psi_\mathrm{c})$, and an application graph as $g_i = (V_{i}, E_{i}, A_i, R_{i}, \phi_{i}, \psi_{i})$. $V_\mathrm{c}$ and $V_{i}$ are the node sets of $G_\mathrm{c}$ and $g_{i}$, respectively; $|V_\mathrm{c}| \gg |V_{i}|$ and $|V_\mathrm{c} \cap V_{i}| / |V_{i}| \lesssim 1$. The node sets $V_\mathrm{c}$ and $V_i$ are mapped by the node-type mapping functions $\phi_\mathrm{c} : V_\mathrm{c} \rightarrow A_\mathrm{c}$ and $\phi_{i} : V_i \rightarrow A_i$, respectively, where $A_\mathrm{c}$ and $A_i$ are the sets of node types, and $A_i \subseteq A_\mathrm{c}$. $E_\mathrm{c}$ and $E_i$ are the edge sets of $G_\mathrm{c}$ and $g_i$, which are mapped by the edge-type mapping functions $\psi_\mathrm{c}: E_\mathrm{c} \rightarrow R_\mathrm{c}$ and $\psi_i : E_i \rightarrow R_i$, respectively. $R_\mathrm{c}$ and $R_i$ are the sets of edge types of $G_\mathrm{c}$ and $g_i$, and they can be different from each other, i.e., $0\leq|R_c \cap R_i|\leq |R_c|$. 

To leverage the large graph corpus $G_\mathrm{c}$, we pre-train a model that consists of an LM (or multiple LMs) and a multi-relational GNN on $G_\mathrm{c}$ using unsupervised learning, and aim to find the optimized model $\Theta_\mathrm{c}$ by minimizing the empirical loss
\begin{equation}
    \Theta_\mathrm{c}^* = \argmin \mathcal{L}_{c} \left(F_\mathrm{c}(\Theta_\mathrm{c}; G_\mathrm{c}) \right) .
\end{equation}
In various scenarios, the number of LMs employed for modeling different types of nodes can vary. For example, previous knowledge shows that utilizing separate LMs to model different entities in e-commerce scenarios is effective \cite{zhu2021textgnn, li2021adsgnn, ioannidis2022efficient}. Herewith, we employ different LMs for query and product entities present in e-commerce data in this work.

The model $\Theta_\mathrm{c}$ is then fine-tuned on multiple tasks applied on $\mathcal{G}$ to improve their performance. For an application graph $g_i$, we initialize its model $\Theta_i$ by $\Theta_\mathrm{c}^*$, and then fine-tune $\Theta_i$ by minimizing its empirical loss
\begin{equation}
    \Theta_\mathrm{i}^* = \argmin \mathcal{L}_{i} \left(F_i (\Theta_i; g_i) \right) .
\end{equation}

\subsection{The Large Graph Corpus}

The large graph corpus is a prevalent yet novel concept in this work. Generally, it is a heterogeneous graph in which nodes preserve text information and different types of edges can connect the same pairs of nodes. A toy example of a large graph corpus is shown in \autoref{fig:coregraph}, that is, a user behavior graph from a shopping engine which includes the common user behaviors (e.g., ``add'', ``click'', ``like'', and ``purchase'') that happen between user-inputs ``query'' and items ``product''. For example, the relation ``add'' between queries and products represents that users add items to their shopping carts under the queries, and the relation ``like'' represents that users show preferences for certain products and are more inclined to buy them. While the user behaviors are between the two node types of ``query'' and ``product'', the large graph corpus can preserve more relation types that can be either between different node types or among the same node type. There are no specific constraints on the graph schema of a large graph corpus.

Regarding what downstream applications can benefit from the pre-training on a large graph corpus, it can be related to the node and relation types of the large graph corpus. From the perspective of node schema, the large graph corpus can support all applications whose task-required node types are a subset of its node types. However, it is more rigorous to consider the extent of relevance between the relation types of the large graph corpus and the relation types of downstream applications.
In principle, it is not necessary for the relation types of applications to exactly be a subset of the relation types of the large graph corpus. As long as the information gathered through the relations in the large graph corpus is useful for the applications, it should be beneficial to pre-train a model on the large graph corpus.
The nodes in the large graph corpus are associated with abundant text information that can be highly complicated, thus, we utilize large LMs to encode the text information. 
Herewith, we expect that the pre-trained LMs with rich relational information on the large graph corpus can generalize to accommodate a variety of downstream applications with different tasks.

\subsection{Multiple Applications}
We employ various applications with different tasks to investigate how the pre-training on a large graph corpus aids in downstream applications. \autoref{fig:appgraph} shows example graphs of several applications. It is usual that applications preserve a small number of relation types (e.g., one or two), and the relation types can be distinct from those of the large graph corpus. For empirical studies, we form a dataset group \amzcore using Amazon internal datasets, which consists of a large graph corpus for pre-training, and three real applications, \searchctr, \esci, and \qpt.

\textbf{\searchctr.} It is an application to predict whether a user/query would click an advertised product or not, which can be modeled as a link prediction task. Its application graph contains node types ``product'' and ``query'', and an edge type ``ads-click'' which is defined as the click-through rate and differs from the common ``click''.

\textbf{\esci.} It is an application for the query-product matching problem. Its graph contains node types ``query'' and ``product'' and an edge type ``match''. Edges are originally labeled as E (exact), S (substitute), C (complement), and I (irrelevant); however, the task in this work is to classify the edges connecting query-product pairs into one of the two classes, E/S/C and I, i.e., edge classification.

\textbf{\qpt.} It is an application to predict the types of products that are mapped to a query. Its graph contains the edge type ``click'' and node types ``query'' and ``product'', and the task of the application is to classify the ``query'' nodes by multiple labels (out of 403 classes), i.e., multi-label node classification. \newline

\begin{figure}
     \centering
         \includegraphics[width=0.9\columnwidth]{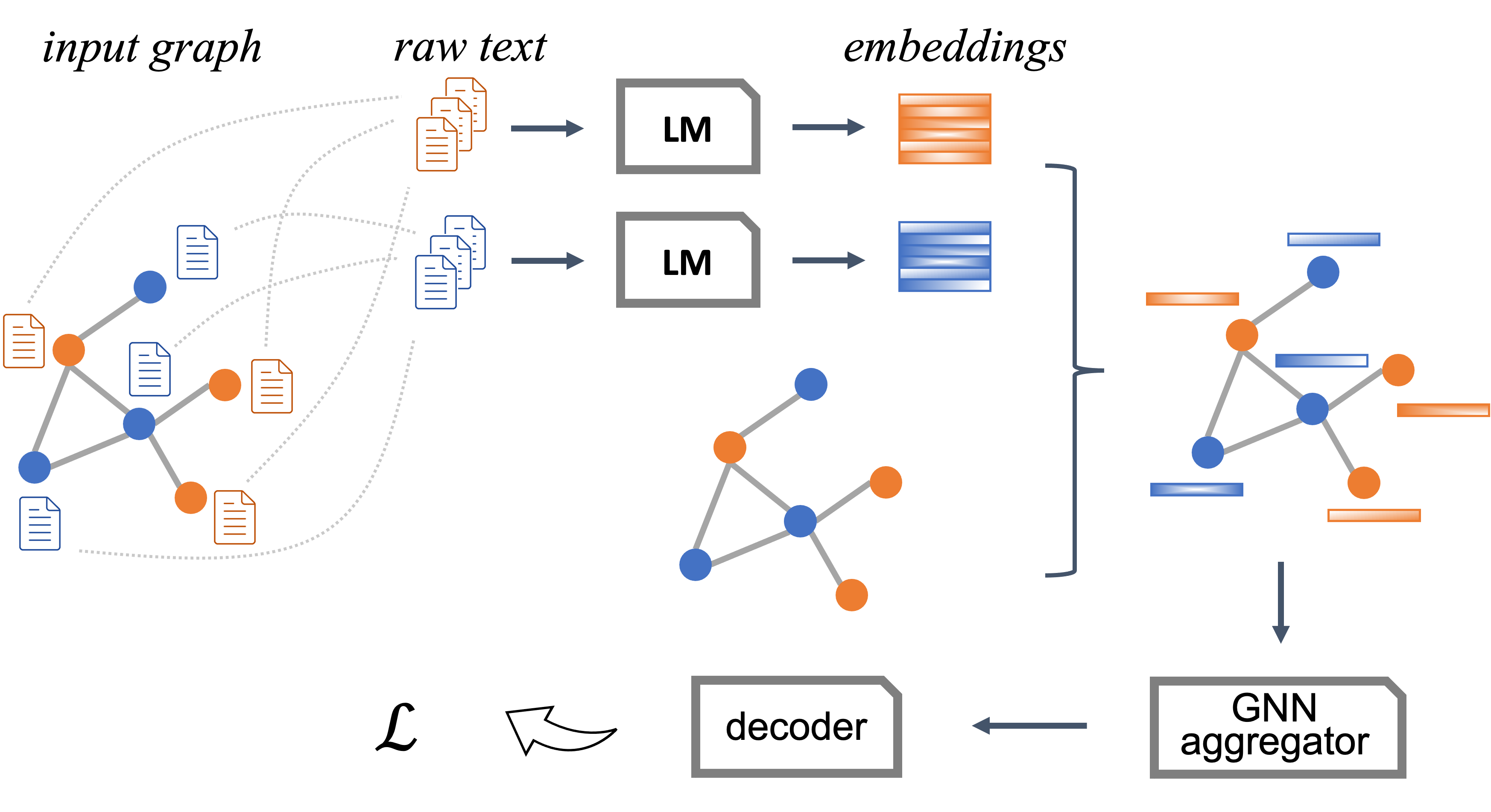}
    \caption{The backbone model \lmgnn. \textmd{\small{Nodes in different colors belong to different node types. The raw text is encoded using an LM or multiple LMs w.r.t. different node types. The output of LMs is then encoded by a GNN aggregator based on the graph topology and finally decoded for a graph-based task.}}} 
\label{fig:lmgnn}
\end{figure}

The data statistics of \amzcore are shown in \autoref{tab:datastats_amzcore} in Appendix \ref{app:datastats_amzcore}. The large graph corpus and application graphs are strategically down-sampled from the original Amazon internal datasets, which are then aggregated and anonymized, and are not representative of real product traffic. The sampling details are discussed in Section \ref{subsubsec:sampling}.

\begin{figure*}
     \centering
     \begin{subfigure}[b]{0.44\linewidth}
         \centering
         \includegraphics[width=\linewidth]{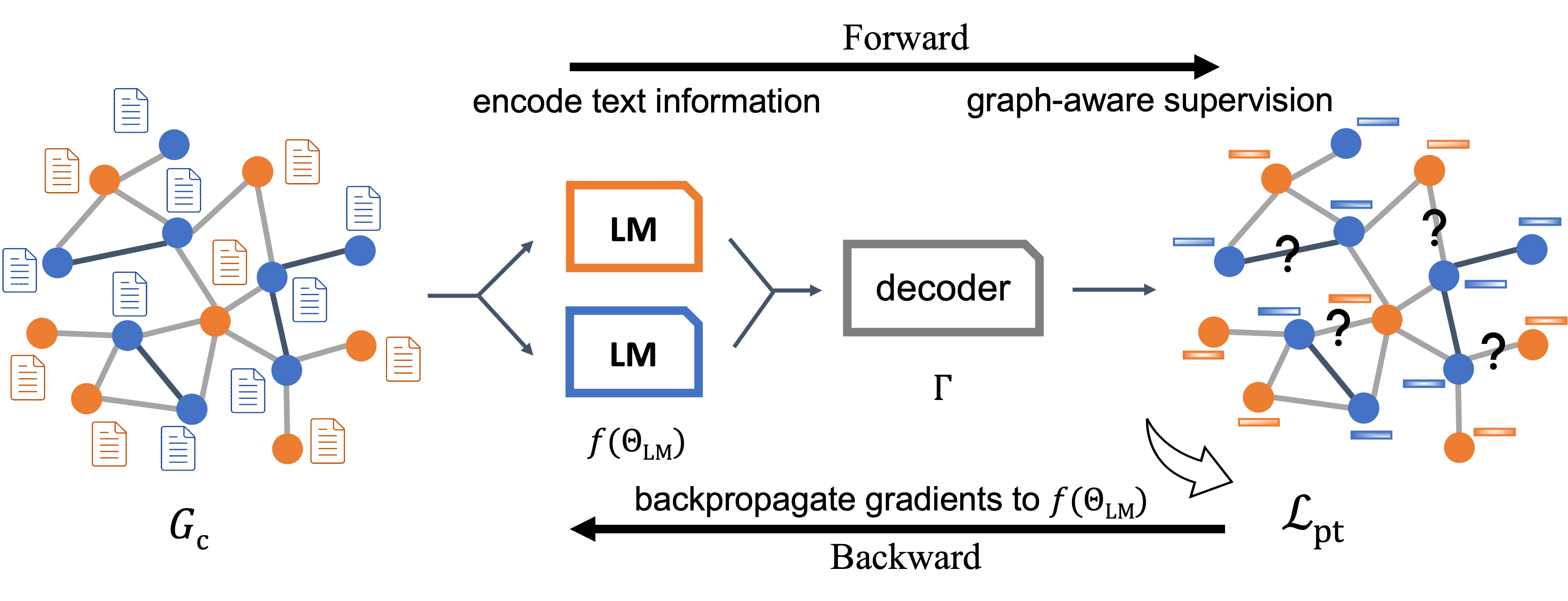}
         \caption{A GNN-free \galm}
         \label{fig:galm}
     \end{subfigure} \hfill
     \begin{subfigure}[b]{0.55\linewidth}
         \centering
         \includegraphics[width=\linewidth]{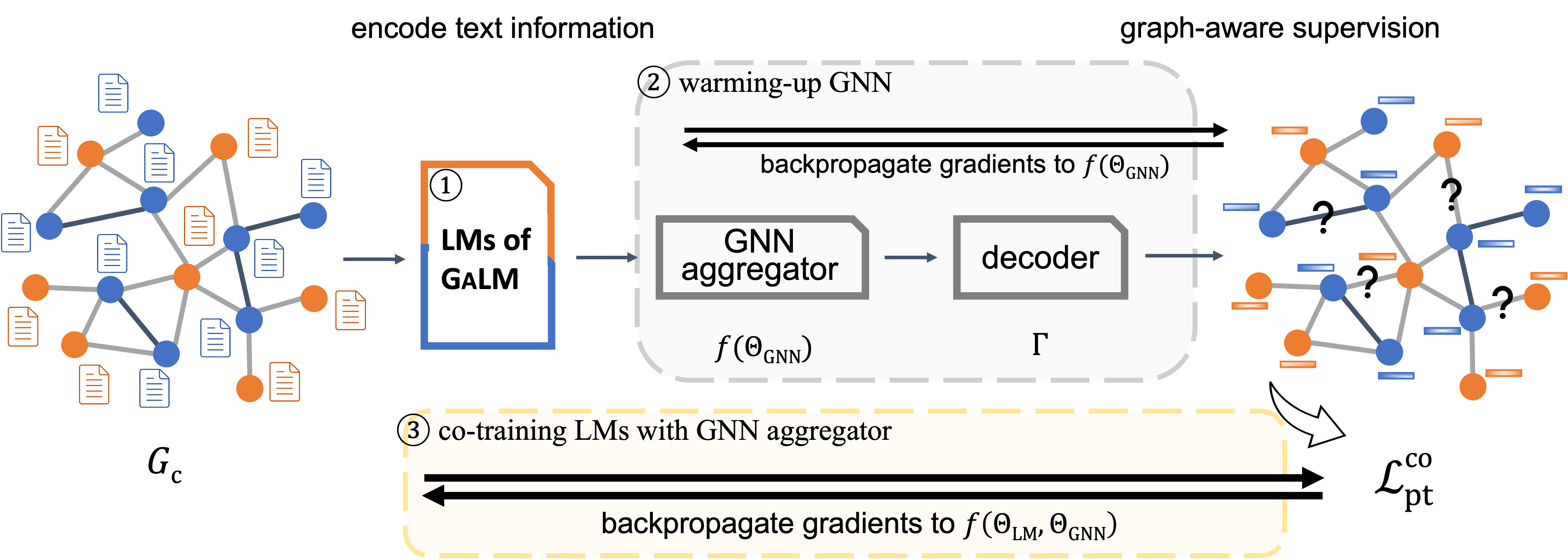}
         \caption{A GNN-co-trained \galm: \galmco}
         \label{fig:galmco}
     \end{subfigure}
    \caption{Graph-aware LM pre-training framework (\galm). \textmd{\small{The LMs are pre-trained on a given large graph corpus either with or without the incorporation of GNN aggregators (\galmco or \galm, respectively). (a) The pre-training of \galm further trains existing general LMs in a graph-aware manner. (b) The pre-training of \galmco includes three steps: i. fine-tuning existing general LMs by graph-aware supervision, ii. warming up the GNN aggregator by fixing the graph-aware pre-trained LMs, iii. co-training the graph-aware pre-trained LMs with the warmed-up GNN aggregator by end-to-end backpropagating the loss to LMs through GNN.}}} 
\label{fig:galms}
\end{figure*}

\section{LM+GNN: The Backbone of Our Proposed Framework}
\label{sec:lmgnn}

The backbone model of our framework for learning on graph corpora is defined as \lmgnn, which consists of one or more LMs for encoding text information and a GNN aggregator for information aggregation.
\autoref{fig:lmgnn} shows the pipeline of \lmgnn. Given a graph corpus as the input, \lmgnn employs one or multiple LMs as text encoder(s) for nodes. Through the LMs, the generated embeddings are then attached to the graph topology as the input of a GNN aggregator, and the output is passed through a task-specific decoder for supervision.

The LMs here can be adapted to various pre-trained large language models. We implement the LMs with BERT in this work because it is well-studied in most existing related works and widely applied in the industry.

The GNN aggregator can also be adapted to different GNN variants, such as GCN and GAT for graphs with only one type of relation, and RGCN and RGAT for graphs with multiple types of relations. The GNN aggregator in the pipeline aids in the propagation of information based on the graph topology. 
Regarding multi-relational graphs, which are more generic and adopted in this work, we implement the GNN aggregators with two common relational GNNs, RGCN \cite{schlichtkrull2018modeling} and RGAT \cite{busbridge2019relational}. The propagation rule of RGCN is
\begin{equation}
    h_u^{(l+1)} = \sigma \left( \sum_{r \in R} \sum_{v \in \mathcal{N}_u^r} \frac{1}{c_{u, r}} \Theta_r^{(l)} h_v^{(l)} + \Theta_0^{(l)} h_u^{(l)} \right),
\end{equation}
where $h_u^{(l)}$ is the representation of node $u$ at the layer $l$, and node $v$ is a neighbor of node $u$ that is connected by an edge with relation $r$. The normalization constant can be customized w.r.t. specific tasks, such as $c_{u, r} = |\mathcal{N}_u^r|$ for entity classification task, and $c_{u, r} = \sum_r |\mathcal{N}_u^r|$ for link prediction task. For RGAT, it introduces the attention mechanism into RGCN and learns the normalized attention coefficients for aggregating neighborhood information with attention.

In the vanilla \lmgnn model, the LMs and a GNN aggregator are simply stacked, and the two processes of encoding text information and aggregating neighborhood information are separate.

\section{Graph-aware LM Pre-training on a Large Graph Corpus}
\label{sec:pretraining}

The pre-training on a large graph corpus aims to capture as much information as possible that can aid in a variety of downstream applications. As LMs retain a significant number of parameters that can understand the text information well, and graph topology conserves the correlation between entities that can be meaningful for propagating information, we propose a \textbf{G}raph-\textbf{A}ware \textbf{LM} pre-training framework (a.k.a. \textbf{\galm}) to combine their advantages. Pre-training \galm on a large graph corpus can incorporate the graph information into the fine-tuning of existing pre-trained LMs that are pre-trained on general texts (such as the BERT model). Although it is in fact fine-tuning existing pre-trained LMs on a large graph corpus, we define it as graph-aware LM pre-training from the perspective of our overall framework. The pipeline of \galm is displayed in \autoref{fig:galms}.

\subsection{Graph-aware LM Pre-training}
\label{subsec:galm}

Graph-aware LM pre-training (\galm) aims to pre-train LMs on a large graph corpus with its graph information being absorbed in the LMs. A straightforward way of incorporating the graph information into LM pre-training is utilizing graph tasks for supervision. 
\autoref{fig:galm} displays the pipeline of a \galm. Given a large graph corpus, its attached raw text information is encoded by the general pre-trained LMs. As separate LMs are usually employed for query and product entities in e-commerce scenarios, we introduce our framework by the way that separate LMs are responsible for different node types of the large graph corpus in the following. The output embeddings of LMs are then associated with the graph topology as the inputs of a graph-task-specific decoder. The full forward pass is supervised by the graph task and computes the loss
\begin{equation}
    \mathcal{L}_\mathrm{pt} = \ell \left( \Gamma \left( \Theta_\mathrm{de} ; f(\Theta_\mathrm{LM}; G_\mathrm{c}) \right) \right),
\end{equation}
where $\Theta_\mathrm{LM}$ and $\Theta_\mathrm{de}$ are the parameters of the LMs and the decoder $\Gamma$ of \galm, and $\ell$ is the task-specific loss. In this work, we employ the typical unsupervised learning task for pre-training on a large graph corpus, i.e., link prediction. Specifically, the loss for a link prediction task is computed by 
\begin{equation}
    \hat{y}_{e_{uv}} = \Gamma (\Theta_\mathrm{de}; h_u \oplus h_v),
\label{eq:catembs}
\end{equation}
\begin{equation}
    \mathcal{L}_\mathrm{pt} = - \sum_{e_{uv} \in E_\mathrm{c}^\mathrm{train}} \left( y_{e_{uv}} \log(\hat{y}_{e_{uv}}) + (1-y_{e_{uv}}) \log(1-\hat{y}_{e_{uv}}) \right),
\label{eq:celoss}
\end{equation}
where $h_u$ and $h_v$ are the output embeddings of the LMs of \galm. In the backward process, the loss $\mathcal{L}_\mathrm{pt}$ of the forward propagation will be backpropagated to the LMs by using the gradients to fine-tune the parameters of LMs.

To investigate whether the graph-aware pre-training on the large corpus is effective in promoting multiple applications, we compare the pre-trained \galm with two baseline models, \hf and \bertmlm, by applying them to several applications with different tasks. The model \hf is initialized from the public BERT model that is pre-trained on English language in the general domain using masked language modeling (MLM) objective, and the model \bertmlm is further trained based on \hf using the text information alone from our large graph corpus (aggregated node features without graph structures) by MLM. From \autoref{tab:bert_pretrain}, it is obvious that \galm significantly improves the applications by comparing to \hf and \bertmlm, which indicates that the graph-aware LM pre-training on a large graph corpus is capable of capturing useful information that could be beneficial to downstream applications. Additionally, the finding demonstrates the key point of \galm that distinguishes our work from the priors, which focuses on how the pre-training, especially, the graph-aware LM pre-training acts in helping with multiple applications.

\begin{table}[htbp]
\centering
\caption{The effect of graph-aware LM pre-training. \textmd{\small{Due to company regulations, the results of \hf baseline are set to zeros, and other baselines and \galm models report relative increases from it.}}}
\resizebox{0.8\columnwidth}{!}{
    \begin{threeparttable}
    \begin{tabular}{ l | c c c }
    \toprule
     \multirow{2}{*}{Models} & \searchctr & \esci & \qpt \\
    & {\footnotesize(ROC-AUC)} & {\footnotesize(macro-F1)} & {\footnotesize(macro-F1)}\\
    \midrule
    \midrule
    \hf & 0.00 & 0.00 & 0.00 \\
    \bertmlm & 0.31\% & -0.24\% & 19.00\% \\
    \midrule
    \bertwg & 2.54\% &  2.78\% & 32.88\% \\
    \bertcorgcn & 3.00\% & 3.55\% & 30.38\% \\
    \bertcorgat & 3.16\% &  2.95\% & 32.80\% \\
    \bottomrule
    \end{tabular}
    \begin{tablenotes}[flushleft]\footnotesize
        \item * The rule is applied to all results evaluated on internal applications of \amzcore due to the legal issue.
    \end{tablenotes}
    \end{threeparttable}
    }
\label{tab:bert_pretrain}
\end{table}

\subsection{GNN-based Graph-aware LM Pre-training}
\label{subsec:galmco}

Another natural alternative to incorporating graph information into LMs is co-training LMs with a GNN aggregator. In addition to the aforementioned way of using graph tasks as the supervision, co-training LMs with a GNN aggregator performs the end-to-end training on pre-trained LMs. However, it is known that the end-to-end training of LMs with GNNs can be extremely time-consuming, especially on a large graph corpus. Thus, we choose to only backpropagate on samples in the co-training process. To accelerate the overall converging speed, we perform two additional steps before co-training by infusing LMs with graph information overhead. We first fine-tune existing general LMs on the large graph corpus by graph-aware supervision. Then, we warm up the GNN aggregator to prevent the co-training from settling on undesirable local minima as the LMs are optimized (also discussed in \cite{ioannidis2022efficient}).

\autoref{fig:galmco} details the pipeline of the GNN-based graph-aware LM pre-training (we term it as \galmco), which includes the training of a vanilla \galm (see Section \ref{subsec:galm}), warming up a GNN aggregator, and co-training the pre-trained LMs of \galm with the warmed-up GNN aggregator. The GNN warming-up step follows \lmgnn (see Section \ref{sec:lmgnn}), which fixes the LMs of a vanilla \galm and trains a GNN aggregator with link prediction. Empirical studies show that a fixed number of epochs (e.g., 2 or 3) is sufficient for warming up. In the co-training step, the forward pass computes the loss
\begin{equation}
    \mathcal{L}_\mathrm{pt}^\mathrm{co} = \ell (\Gamma ( \Theta_\mathrm{de}; f (\Theta_\mathrm{LM}, \Theta_\mathrm{GNN}; G_\mathrm{c}))),
\end{equation}
and the backward pass will backpropagate the gradients on some samples from the GNN aggregator to the LMs.

To study whether the GNN-based graph-aware LM pre-training will further promote the applications, we adopt two types of GNN aggregators into \galmco, RGCN \cite{schlichtkrull2018modeling} and RGAT \cite{busbridge2019relational}, and compare the performance of the LMs from \galmco (\bertcorgcn and \bertcorgat) to those from the vanilla \galm on applications, as shown in \autoref{tab:bert_pretrain}. In general, \galmco performs similarly to the vanilla \galm on the applications. We observe a slight improvement of \galmco on \searchctr, but it does not show superiority on \esci and \qpt. Since \galmco is comparable to a vanilla \galm on downstream applications but is more time-consuming (see Appendix \ref{app:pretraining_time}), we deem it more reasonable to choose the vanilla graph-aware LM pre-training method rather than the GNN-based one for pre-training on a large graph corpus, considering the restrictions of computation and time resources.

\section{\galm Fine-tuning on Multiple Graph Applications}
\label{sec:finetuning}
\begin{figure}
     \centering
         \includegraphics[width=\columnwidth]{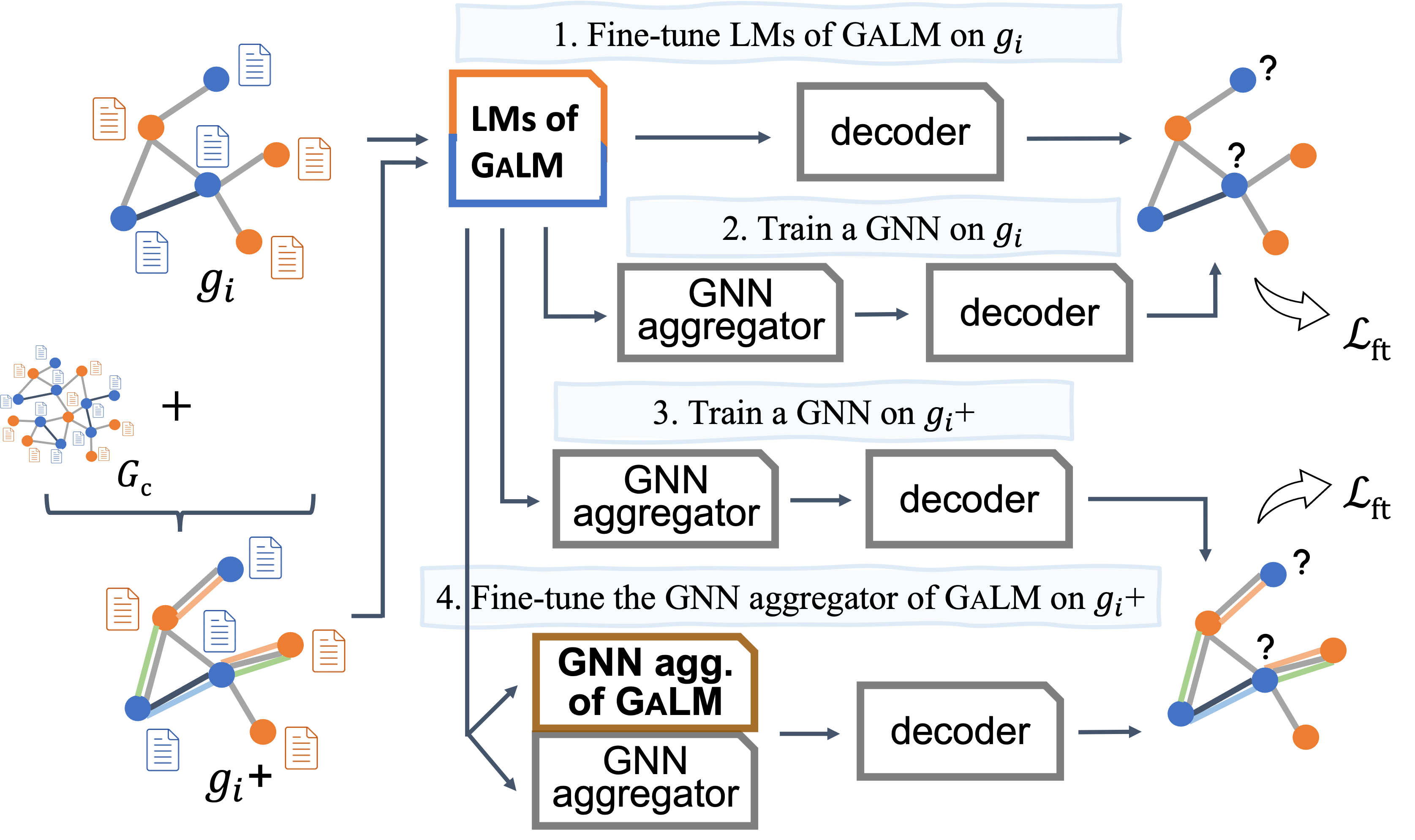}
    \caption{Fine-tuning strategies. \textmd{\small{Step 1: Graph-aware LM fine-tuning on applications using the pre-trained \galm; Step 2: Fine-tuning \galm (or \bertwgft) on an application by training a GNN aggregator; Step 3: Stitching an application graph with the large graph and training a GNN aggregator on the stitched application graph; Step 4: Fine-tuning the GNN aggregator of \galm on a stitched application graph.}}} 
\label{fig:finetune}
\end{figure}

We demonstrated the effectiveness of \galm in Section \ref{sec:pretraining} by directly applying its graph-aware-pre-trained LMs to various applications. In this section, furthermore, we investigate how the \galm can be fine-tuned to further improve the performance on multiple graph applications. As the architecture of a \galm consists of (node-type-specific) LMs, an optional relation-type-specific GNN aggregator, and a task-specific decoder, the fine-tuning of a \galm can involve fine-tuning LMs of \galm, along with training or fine-tuning a GNN aggregator. The proposed fine-tuning methods of \galm are illustrated in \autoref{fig:finetune}.

\subsection{Fine-tuning \galm on Applications}
To fine-tune the pre-trained \galm on applications, the proposed methods can be 1) further fine-tuning the LMs of \galm on graph applications, and 2) training a new GNN aggregator on application graphs employing the LMs of \galm. 

\subsubsection{\textbf{Graph-aware LM fine-tuning on applications}}
\label{subsubsec:ftlm}

Similar to Section \ref{subsec:galm}, we adopt the graph information into the LM fine-tuning on applications (Step 1 in \autoref{fig:finetune}). Given an application graph with raw text, we use the LMs from a pre-trained \galm to encode the raw text, whose output embeddings are then input into an application-specific decoder. The forward pass is supervised by a particular graph task that varies depending on applications. For example, for an application graph $g_i$ with a node classification task, the loss of its graph-aware LM fine-tuning is computed by 
\begin{equation}
    \mathcal{L}_\mathrm{ft} = \sum_{u \in V_i^\mathrm{train}} - \log \left( \softmax (y_u, \hat{y}_u) \right);
\end{equation}
for an application with an edge classification task, its loss can be computed using the \autoref{eq:catembs} and \ref{eq:celoss}.

To study how the graph-aware fine-tuning of LMs from \galm helps various applications, we fine-tune the pre-trained \galm on applications of \amzcore with diverse graph tasks. The fine-tuned model is denoted as \bertwgft. As can be observed from \autoref{tab:galm_ft}, \bertwgft slightly improves \bertwg on \searchctr and \esci, but significantly outperforms \bertwg on \qpt. The reason can be that \searchctr and \esci conduct similar link-level tasks to \galm during its pre-training on a large graph corpus, while \qpt conducts a multi-label node classification task that differs more from \galm's pre-training task. Hence, \searchctr and \esci can barely gain extra information from fine-tuning LMs of \galm on the application graph, while \qpt can still obtain beneficial new information.

Additionally, although we have shown the direct benefit of pre-training LMs on a large graph corpus without fine-tuning on the application graph (\autoref{tab:bert_pretrain}), it might be interesting to see what if the generic LMs are directly fine-tuned on the application graph. To this end, we perform graph-aware LM fine-tuning using the public BERT model on these applications, as denoted by \hfft in \autoref{tab:galm_ft}. By comparing \bertwgft and \hfft, it is evident that the pre-training on the large graph corpus can benefit the applications to varying extents depending on the applications.

\begin{table}[htbp]
\centering
\caption{The effect of fine-tuning \galm on applications.}
\resizebox{0.8\columnwidth}{!}{
    \begin{tabular}{ l | c c c }
    \toprule
     \multirow{2}{*}{Models} & \searchctr & \esci & \qpt \\
    & {\footnotesize(ROC-AUC)} & {\footnotesize(macro-F1)} & {\footnotesize(macro-F1)} \\
    \midrule
    \midrule
    \hfft & 1.58\% & 2.92\% & 4.89\% \\
    \bertwg & 2.54\% &  2.78\% & 32.88\% \\
    \midrule
     \bertwgft & 3.04\% & 3.89\% & 41.20\% \\
     \galmrgcnft & 4.46\% & 4.96\%  & 61.06\%\\
     \galmrgatft & 17.49\% &  10.89\% & 66.50\% \\
    \bottomrule
    \end{tabular}}
\label{tab:galm_ft}
\end{table}

\subsubsection{\textbf{Fine-tuning \galm with GNNs on applications}}
\label{subsubsec:galmgnn}

As the graph-aware LM fine-tuning on applications can encode the graph information into LMs to some extent, it is interesting to explore whether leveraging an extra GNN aggregator to propagate information is still desirable. We regard this method of training application-specific GNN aggregators using LMs of \bertwgft as fine-tuning \galm with GNNs on applications.
In principle, the fine-tuning can be realized by training a standalone GNN aggregator using the LMs of \bertwgft or co-training a GNN aggregator with the LMs. Considering the trade-offs between utility and efficiency, it is reasonable to choose the former (Step 2 in \autoref{fig:finetune}), where we use the LMs of \bertwgft to generate embeddings for nodes of an application graph, and train a GNN aggregator and a task-specific decoder on the application graph with the generated embeddings as input features.

To investigate the effect of fine-tuning \galm with GNNs on applications, we further fine-tune a \bertwgft with RGCN and RGAT on the internal applications, as denoted by \galmrgcnft and \galmrgatft, respectively. From \autoref{tab:galm_ft}, the improvements brought by \galmrgcnft and \galmrgatft are significant compared to \bertwgft, which indicates that fine-tuning \bertwgft by training GNN aggregators on applications is profitable.

\subsection{Fine-tuning \galm on Applications Stitched with the Large Graph Corpus}

Apart from fine-tuning \galm with GNNs on applications at the model level, another way to inject graph information is by stitching application graphs with the large graph corpus at the data level (we term the resulting graphs as stitched application graphs, $\{g_i+\}$). Since the large graph corpus preserves more edge types and many more edges that are not included in applications, stitching application graphs with the large graph would introduce more neighbors for nodes in applications, which could benefit their gathering of neighborhood information. 

To stitch an application graph with the large graph corpus, we align the overlapped nodes between the application graph and the large graph via internal entity IDs, and then add the connecting edges of these overlapped nodes that appear in the large graph into the application graph.

\subsubsection{\textbf{Fine-tuning \galm with GNNs on stitched application graphs}}
\label{subsubsec:galmgnncore}

Similar to Section \ref{subsubsec:galmgnn}, fine-tuning \galm with GNNs on a stitched application graph aims to train a GNN aggregator with an application-specific decoder on the stitched application graph utilizing pre-trained LMs of \galm (Step 3 in \autoref{fig:finetune}). Here, we employ LMs of \bertwgft to generate node embeddings and train a GNN aggregator and decoder on the stitched application graph; the fine-tuned models are \galmrgcnftplus and \galmrgatftplus, respectively.

From \autoref{tab:coreMP}, both \galmrgcnftplus and \galmrgatftplus are apparently superior to \galmrgcnft and \galmrgatft, respectively, which indicates that applications can benefit from introducing more neighborhood information from the large graph corpus. On the other hand, the superiority of \galmrgnnftplus over \galmrgnnft implies that the model-level graph-aware LM pre-training on the large graph corpus using link prediction is insufficient for capturing all the graph information from the large graph corpus. Consequently, enhancing the training of GNN aggregators at data level on the stitched application graphs with extra edges from the large graph using \bertwgft can further improve the performance of applications. However, it remains an open question whether the graph information of the large graph corpus can be fully captured through more dedicated designs of LMs and graph-aware pre-training, which could be explored in future work.

\begin{table}[htbp]
\centering
\caption{The effect of fine-tuning \bertwgft with GNNs on stitched application graphs.}
\resizebox{0.8\columnwidth}{!}{
    \begin{tabular}{ l | c c c }
    \toprule
      \multirow{2}{*}{Models} & \searchctr & \esci & \qpt \\
     & {\footnotesize(ROC-AUC)} & {\footnotesize(macro-F1)} & {\footnotesize(macro-F1)} \\
    \midrule
    \midrule
    \galmrgcnft & 4.46\% & 4.96\% & 61.06\% \\
    \galmrgatft & 17.49\% & 10.89\% & 66.50\% \\
    \midrule
     \galmrgcnftplus & 13.38\% &  8.19\% & 76.39\% \\
     \galmrgatftplus & 20.37\% & 13.35\% & 80.06\% \\
    \bottomrule
    \end{tabular}}
\label{tab:coreMP}
\end{table}

\subsubsection{\textbf{Fine-tuning \galm with pre-trained GNNs on stitched application graphs}}
\label{subsubsec:ftgnn}

As discussed in Section \ref{subsec:galmco}, in the stage of pre-training on a large graph corpus, besides LMs, \galm can also pre-train a standalone GNN aggregator or co-training a GNN aggregator with the LMs. Thus, it is natural to investigate whether the pre-trained GNN aggregator can be fine-tuned on applications, especially under the consideration of training efficiency. However, due to the inconsistency of edge types between the applications and the large graph, the pre-trained GNN aggregator that models the specific types of edges in the large graph cannot be directly fine-tuned on the application graphs. A proposed method is to first include the original edge distribution in applications, that is, to stitch application graphs with the large graph; then, in addition to inheriting the pre-trained GNN aggregator for the original types of edges in the large graph, we initialize a new GNN aggregator for the new types of edges in the application graphs (Step 4 in \autoref{fig:finetune}).

The proposed method of fine-tuning the pre-trained GNN aggregator on stitched application graphs seems to be straightforward. However, it in fact involves complex decision-making-- for example, how to choose a combination of the learning rates for fine-tuning the pre-trained GNN aggregator and training the newly initialized GNN aggregator, which would affect how much information from the pre-training to be retained; how to initialize the new GNN aggregator, where different initializations may impact the optimizing; and how to aggregate the outputs of the pre-trained and the newly initialized GNN aggregators, which can be a simple operation (e.g., summation, average, and concatenation) or parameterized and to be learned. We deem these complicated decisions beyond the scope of this work and will only provide preliminary results and leave the further explorations in future work.

Albeit the complex decision-making, we attempted some simple decisions to fine-tune the pre-trained GNN aggregator using \bertwgft, i.e., initializing the new GNN aggregator at Xavier's random \cite{pmlr-v9-glorot10a}, summing up the outputs of the pre-trained and new initialized GNN aggregators for the same nodes, and tuning the learning rate of the GNN fine-tuning to tens, hundreds, or thousands of times smaller than the learning rate of the new GNN aggregator. The fine-tuned model \galmrgatftplusft in \autoref{tab:gnnft} displays the best results according to our limited attempts based on the aforementioned simple decisions. In general, the \galmrgatftplusft model is found comparable but slightly inferior to \galmrgatftplus on the stitched application graphs. 
On the other hand, we observe that fine-tuning the pre-trained GNN aggregator of \galm on stitched application graphs may indeed lead to faster convergence. For example, \galmrgatftplusft converges to 97\% of the best performance on \searchctr after $600$ iterations while \galmrgatftplus needs $13,800$ iterations to converge to the same level, where the elapsed time of every iteration is roughly the same across the two frameworks.

\begin{table}[htbp]
\centering
\caption{The effect of fine-tuning \bertwgft with pre-trained GNNs on stitched application graphs.}
\resizebox{0.8\columnwidth}{!}{
    \begin{tabular}{ l | c c c }
    \toprule
      \multirow{2}{*}{Models} & \searchctr & \esci & \qpt \\
     & {\footnotesize(ROC-AUC)} & {\footnotesize(macro-F1)} & {\footnotesize(macro-F1)} \\
    \midrule
    \midrule
     \galmrgatftplus & 20.37\% & 12.99\% & 80.06\% \\
    \midrule
      \galmrgatftplusft & 19.93\%  &  11.34\%  & 77.25\%\\
    \bottomrule
    \end{tabular}}
\label{tab:gnnft}
\end{table}

\section{External and Overall Evaluations}
\label{sec:exprs}

\subsection{Public Data Source}

To solidify our pre-training and fine-tuning methods, apart from the Amazon internal dataset \amzcore, we simulate a large graph corpus and corresponding application graphs using a public dataset, Amazon Product Reviews\footnote{\url{https://jmcauley.ucsd.edu/data/amazon/}} with the time span from May 1996 to July 2014, which we name them together as \amzreviews dataset. The original Amazon Product Reviews dataset is large, so we select some fields of text and down-sample records of items and reviews.

\subsubsection{\textbf{Data partitioning}}
Upon the enormous raw text, we concatenate the selected fields ``title'', ``brand'', ``feature'', and ``description'', as the raw text of items, and select the field ``reviewText'' as the raw text of reviews. We also retain the field ``categories'' as the targets of items for an application task of predicting the product type of items.

The filtered text is then partitioned into a large graph corpus and multiple applications, based on their node and edge schemas. For \amzreviews, the node schema is defined as \{``asin'', ``reviewText''\}. In the large graph corpus, the edge schema is \{``asin--coview--asin'', ``reviewText--review--asin'', ``reviewText--cowrite--reviewText''\}, which separately represent that a reviewText reviews an item, two reviewTexts are written by the same reviewer, and an item is also viewed when its connected item is reviewed. We simulate two application graphs according to the real-world applications, \copurchase and \acat. The new edge type in \copurchase and \acat is ``asin--cobuy--asin'', which represents that a user bought both the connected two items. Herewith, the application of \copurchase is to predict whether users would buy two items together, which performs as a link prediction task; and the \acat application is to predict the product type of items, which performs as a node classification task.

\subsubsection{\textbf{Graph down-sampling}}
\label{subsubsec:sampling}
Instead of sampling records directly from the filtered text, which could lead to largely disturbed graph structures and extremely sparse edges, we first construct the large graph and application graphs using all filtered text and then down-sample the constructed graphs. The sampling begins with a random set of nodes and samples their neighbors; the sampled neighbors are then used as the starting set of nodes for sampling neighbors one hop further. After iterating the process several times, we extract all edges from the initially constructed graph whose two-end nodes are within the set of sampled nodes; hereby, the edges that are connected by the sampled nodes but are not sampled during the iteration of sampling will be also included in the sampled graph, to avoid generating a collection of disconnected k-hop networks. The data statistics of the sampled large graph corpus and application graphs are shown in \autoref{tab:datastats_amareviews} in Appendix \ref{app:datastats_amzreviews}.

\subsection{Implementation Details}

As the original Amazon internal datasets and the public Amazon Product Reviews dataset are both very large, their preprocessing including down-sampling is implemented on distributed Spark framework. The experiments use a cluster of 64 Nvidia Tesla V100 32GB GPUs. The hyper-parameter settings for models and experiments are shown in \autoref{tab:hyperparas} in Appendix \ref{app:hyperparas}.

\subsection{Results on Internal and Public Datasets}

\begin{table}[htbp]
\centering
\caption{The overall comparison of models using \galmrgat on \amzcore and \amzreviews. \textmd{\small{The LMs in models without a ``*'' are not fine-tuned on applications.}}}
\resizebox{\linewidth}{!}{
    \begin{tabular}{ l | c c c | c c c }
    \toprule
    \multirow{3}{*}{Models} & \multicolumn{3}{c|}{\amzcore} & \multicolumn{2}{c}{\amzreviews} \\
      & \searchctr & \esci & \qpt & \copurchase & \acat \\
     & {\footnotesize(ROC-AUC)} & {\footnotesize(macro-F1)} & {\footnotesize(macro-F1)} & {\footnotesize(MRR)} & {\footnotesize(macro-F1)} \\
    \midrule
    \midrule
    \hf+\rgat & 15.60\% & 8.51\% & 39.56\% & 0.1941 & 0.7038 \\
    \hfft+\rgat & 15.47\% & 6.49\% & 44.84\% & 0.3230 & 0.7479 \\
    \midrule
    \galmrgat & 17.30\% & 10.39\% & 64.03\% & 0.3461 & 0.7317 \\
    \galmrgatft & 17.49\% & 10.89\%  & 66.50\% & 0.3474 & 0.7491 \\
    \galmrgatftplus & \textbf{20.37\%} & \textbf{13.35\%}  & \textbf{80.06\%} & \textbf{0.4542} & \textbf{0.7813} \\
    \bottomrule
    \end{tabular}}
\label{tab:overall}
\end{table}

From the previous observations, fine-tuning \galm with GNNs can dramatically improve the performance of applications. Thus, we select the better model of \galmrgat and implement different fine-tuning methods upon it, to obtain an overall comparison on both internal \amzcore and public \amzreviews datasets (see \autoref{tab:overall}). The baselines are implemented as the backbone \lmgnn using public pre-trained LMs, and all the GNN aggregators incorporate RGAT. The models \galmrgatft and \galmrgatftplus are as described in Section \ref{sec:finetuning}, which fine-tune the pre-trained LMs of \galm on applications in a graph-aware manner and then train GNN aggregators on application graphs and stitched application graphs, respectively, while \galmrgat uses the pre-trained LMs of \galm without further fine-tuning on application graphs.

From the results, it further confirms that the graph-aware LM pre-training on a large graph corpus and fine-tuning on applications can be beneficial to these applications with various tasks, as can be observed where most \galm variants perform significantly better on all five applications in comparison to the \hf variants. 
Besides, the superiority of \galmrgatftplus over \galmrgatft demonstrates the effectiveness of introducing more neighborhood information of the large graph corpus to applications at the data level, which can assist the information propagation of GNN aggregators on applications with the extra information that is not fully captured by the model-level graph-aware pre-training. 
It is also noticeable that when fine-tuning \galm with GNN aggregators on applications, it could be unnecessary to directly fine-tune the LMs of \galm on application graphs, as can be observed where \galmrgatft performs similarly to \galmrgat. The reason could be that the capacity of GNN aggregators is sufficient for capturing essential information of the application graphs, thus, training GNN aggregators using \bertwgft does not provide much additional gain over \galm.

\subsection{Additional Analysis}

\subsubsection{\textbf{Significance test}}

\begin{figure}
    \begin{subfigure}[b]{0.5\linewidth}
        \centering
         \includegraphics[width=\linewidth]{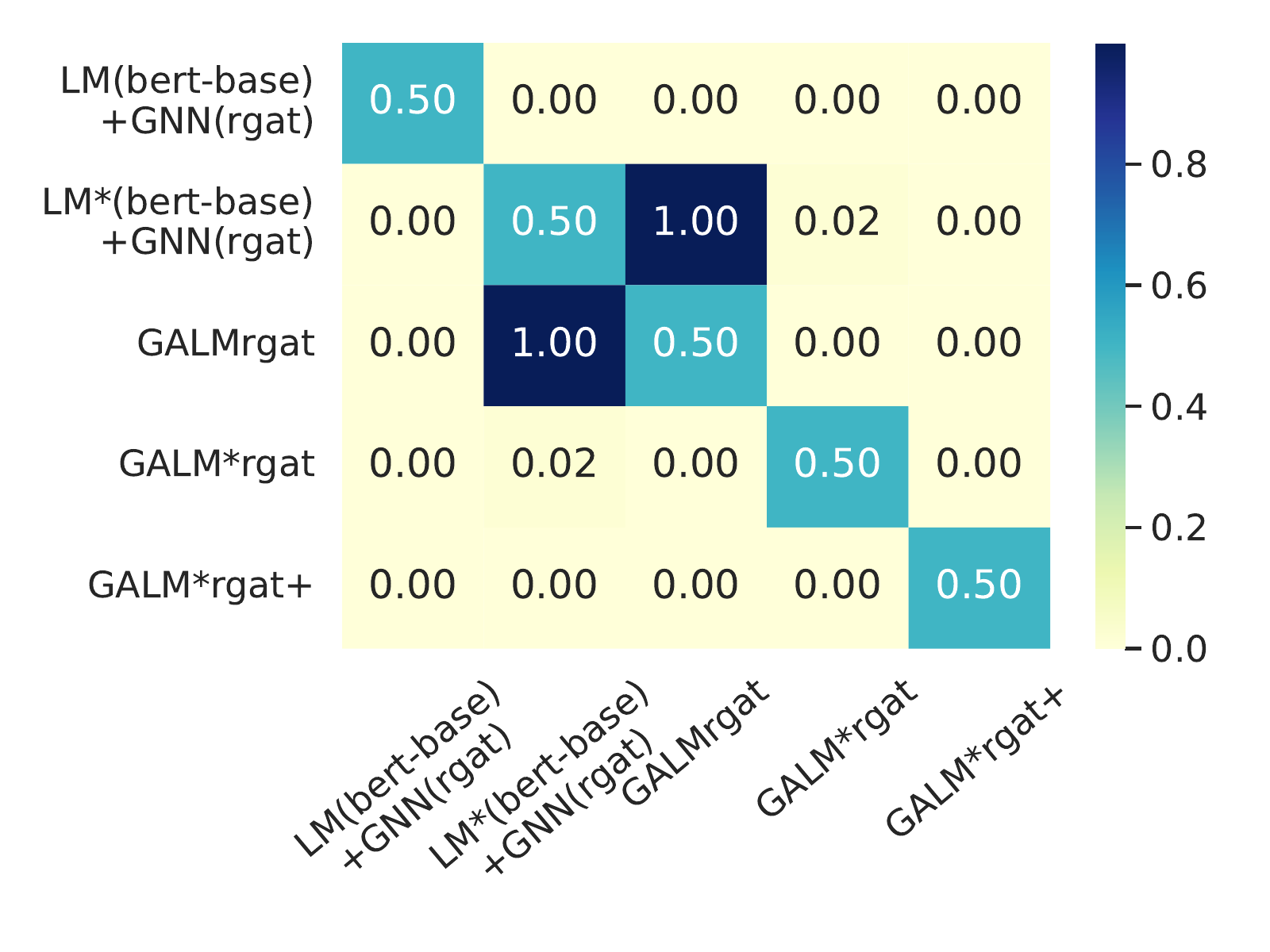}
         \caption{The p-values on \acat.}
        \label{fig:significance}
    \end{subfigure}
    \hfill
     \begin{subfigure}[b]{0.475\linewidth}
     \centering
         \includegraphics[width=\linewidth]{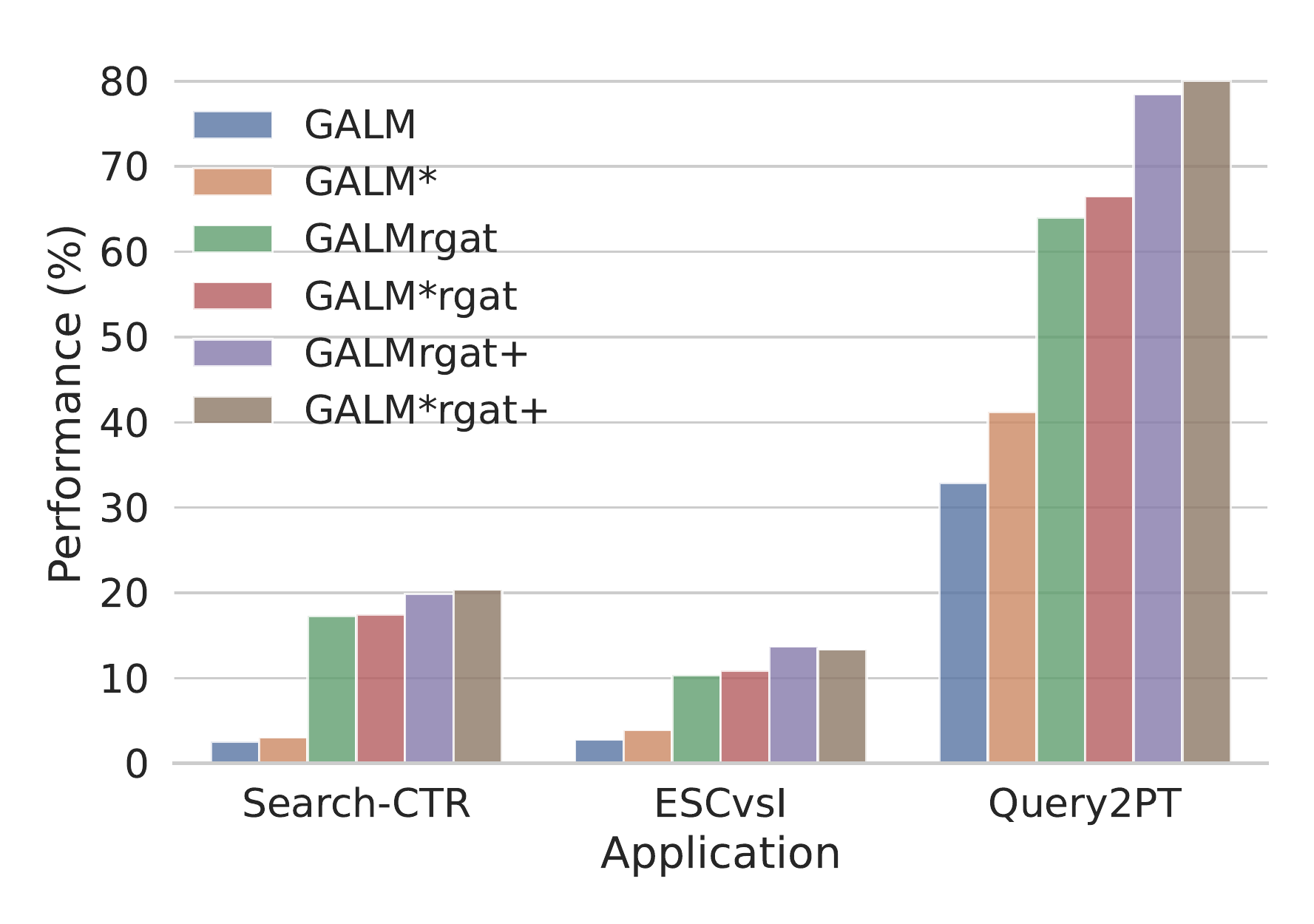}
    \caption{Model performance on \amzcore applications.} 
    \label{fig:ablation}
    \end{subfigure}
\caption{Additional analysis. \textmd{\small{(a) \textit{Significant test.} The p-values are calculated using one-sided pair-wise T-tests (the alternative hypothesis is that the model performs worse than the model below it or on the right side of it). (b) \textit{Ablation studies.} The metrics are ROC-AUC for \searchctr, and macro-F1 for \esci and \qpt.}}}
\label{fig:analysis}
\end{figure}

Regarding the overall comparison in \autoref{tab:overall}, we conduct a significance analysis to confirm the significance of the superiority of our proposed frameworks compared to the baselines. We run five repetitions on the public application \acat for each model (the original results are in \autoref{tab:siganalysis} in Appendix \ref{app:sigresults}), and then perform a one-sided T-test \cite{student1908probable} between each pair of models. By one-sided T-test, we test whether one model performs significantly better than another model rather than only testing whether they are significantly different. \autoref{fig:significance} displays the heatmap of p-values from the pair-wise one-sided T-tests. According to the heatmap, most of the comparison is significant as the p-value is less than 0.05. A noticeable result is the p-value of 1.0 between \hfft+\rgat and \galmrgat, which indicates that the alternative hypothesis that \galmrgat is significantly better than \hfft+\rgat is rejected. It can be because the \acat dataset is relatively small and simple (with a single type of nodes and edges), and the fine-tuning of public pre-trained LMs on it combing with training a GNN aggregator is capable to capture adequate information for its task, so \hfft+\rgat can achieve a comparable result to the \galmrgatft model.

\subsubsection{\textbf{Ablation studies}}

To investigate the effects of different fine-tuning methods with the pre-trained \galm, we discard one fine-tuning strategy from \galmrgatftplus each time and compare the resulting models. The results are visualized in \autoref{fig:ablation}.
It can be seen that fine-tuning LMs of \galm on applications (with a superscript *) has an effect when only LMs are employed (blue and yellow bars). When adopting the strategy of training a GNN aggregator (the four bars on the right), the effect of fine-tuning LMs on applications can be diminished (green v.s. red, and purple v.s. brown). Furthermore, integrating GNN aggregators in the fine-tuning stage can considerably improve the performance of applications, and its effect is greater than the effect of fine-tuning LMs on applications (as can be seen in the figure where the green bar rises steeply from the yellow bar, and the gap is more prominent than the one between the blue and yellow bars). Regarding fine-tuning \galm on stitched application graphs, as it will benefit from employing GNN aggregators with more neighborhood information, it can further promote the performance of applications compared to fine-tuning \galm with GNNs on sole applications themselves.

\section{Conclusion}
The work studies how to leverage a large graph corpus to facilitate multiple downstream graph applications whose edge schemas can be distinct and tasks can vary. To approach this problem, we propose a graph-aware language model pre-training framework \galm and advanced variants by incorporating more fine-tuning strategies. The extensive experiments demonstrate the effectiveness of our proposed framework and fine-tuning methods. Moreover, we provide insights into the empirical results and additional analysis.
A limitation of the work is that we only experiment with the language model of BERT and have not attempted other powerful LMs, which is mainly due to the currently more developed large-scale training pipelines based on BERT in the industry. Besides, we also elicit two open questions in the paper: whether more delicate designs of the pre-training task and language models can fully capture the large graph corpus information, and how to resolve the complex decision-making in fine-tuning \galm with pre-trained GNNs. These can all lead to profitable future studies.




\bibliographystyle{ACM-Reference-Format}
\balance
\bibliography{bibliography}


\begin{thebibliography}{44}


\ifx \showCODEN    \undefined \def \showCODEN     #1{\unskip}     \fi
\ifx \showDOI      \undefined \def \showDOI       #1{#1}\fi
\ifx \showISBNx    \undefined \def \showISBNx     #1{\unskip}     \fi
\ifx \showISBNxiii \undefined \def \showISBNxiii  #1{\unskip}     \fi
\ifx \showISSN     \undefined \def \showISSN      #1{\unskip}     \fi
\ifx \showLCCN     \undefined \def \showLCCN      #1{\unskip}     \fi
\ifx \shownote     \undefined \def \shownote      #1{#1}          \fi
\ifx \showarticletitle \undefined \def \showarticletitle #1{#1}   \fi
\ifx \showURL      \undefined \def \showURL       {\relax}        \fi
\providecommand\bibfield[2]{#2}
\providecommand\bibinfo[2]{#2}
\providecommand\natexlab[1]{#1}
\providecommand\showeprint[2][]{arXiv:#2}

\bibitem[Brown et~al\mbox{.}(2020)]%
        {brown2020language}
\bibfield{author}{\bibinfo{person}{Tom Brown}, \bibinfo{person}{Benjamin Mann},
  \bibinfo{person}{Nick Ryder}, \bibinfo{person}{Melanie Subbiah},
  \bibinfo{person}{Jared~D Kaplan}, \bibinfo{person}{Prafulla Dhariwal},
  \bibinfo{person}{Arvind Neelakantan}, \bibinfo{person}{Pranav Shyam},
  \bibinfo{person}{Girish Sastry}, \bibinfo{person}{Amanda Askell},
  {et~al\mbox{.}}} \bibinfo{year}{2020}\natexlab{}.
\newblock \showarticletitle{Language models are few-shot learners}. In
  \bibinfo{booktitle}{\emph{Advances in neural information processing
  systems}}.
\newblock


\bibitem[Busbridge et~al\mbox{.}(2019)]%
        {busbridge2019relational}
\bibfield{author}{\bibinfo{person}{Dan Busbridge}, \bibinfo{person}{Dane
  Sherburn}, \bibinfo{person}{Pietro Cavallo}, {and} \bibinfo{person}{Nils~Y
  Hammerla}.} \bibinfo{year}{2019}\natexlab{}.
\newblock \showarticletitle{Relational graph attention networks}.
\newblock \bibinfo{journal}{\emph{arXiv preprint arXiv:1904.05811}}
  (\bibinfo{year}{2019}).
\newblock


\bibitem[Chien et~al\mbox{.}(2022)]%
        {chiennode}
\bibfield{author}{\bibinfo{person}{Eli Chien}, \bibinfo{person}{Wei-Cheng
  Chang}, \bibinfo{person}{Cho-Jui Hsieh}, \bibinfo{person}{Hsiang-Fu Yu},
  \bibinfo{person}{Jiong Zhang}, \bibinfo{person}{Olgica Milenkovic}, {and}
  \bibinfo{person}{Inderjit~S Dhillon}.} \bibinfo{year}{2022}\natexlab{}.
\newblock \showarticletitle{Node Feature Extraction by Self-Supervised
  Multi-scale Neighborhood Prediction}. In
  \bibinfo{booktitle}{\emph{International Conference on Learning
  Representations}}.
\newblock


\bibitem[Devlin et~al\mbox{.}(2018)]%
        {devlin2018bert}
\bibfield{author}{\bibinfo{person}{Jacob Devlin}, \bibinfo{person}{Ming-Wei
  Chang}, \bibinfo{person}{Kenton Lee}, {and} \bibinfo{person}{Kristina
  Toutanova}.} \bibinfo{year}{2018}\natexlab{}.
\newblock \showarticletitle{Bert: Pre-training of deep bidirectional
  transformers for language understanding}. In \bibinfo{booktitle}{\emph{arXiv
  preprint arXiv:1810.04805}}.
\newblock


\bibitem[Fang et~al\mbox{.}(2020)]%
        {fang2020pre}
\bibfield{author}{\bibinfo{person}{Yang Fang}, \bibinfo{person}{Xiang Zhao},
  \bibinfo{person}{Yifan Chen}, \bibinfo{person}{Weidong Xiao}, {and}
  \bibinfo{person}{Maarten de Rijke}.} \bibinfo{year}{2020}\natexlab{}.
\newblock \showarticletitle{Pre-Trained Models for Heterogeneous Information
  Networks}.
\newblock \bibinfo{journal}{\emph{arXiv preprint arXiv:2007.03184}}
  (\bibinfo{year}{2020}).
\newblock


\bibitem[Glorot and Bengio(2010)]%
        {pmlr-v9-glorot10a}
\bibfield{author}{\bibinfo{person}{Xavier Glorot} {and} \bibinfo{person}{Yoshua
  Bengio}.} \bibinfo{year}{2010}\natexlab{}.
\newblock \showarticletitle{Understanding the difficulty of training deep
  feedforward neural networks}. In \bibinfo{booktitle}{\emph{Proceedings of the
  Thirteenth International Conference on Artificial Intelligence and
  Statistics}}.
\newblock


\bibitem[Hamilton et~al\mbox{.}(2017)]%
        {hamilton2017inductive}
\bibfield{author}{\bibinfo{person}{Will Hamilton}, \bibinfo{person}{Zhitao
  Ying}, {and} \bibinfo{person}{Jure Leskovec}.}
  \bibinfo{year}{2017}\natexlab{}.
\newblock \showarticletitle{Inductive representation learning on large graphs}.
  In \bibinfo{booktitle}{\emph{Advances in neural information processing
  systems}}.
\newblock


\bibitem[Han et~al\mbox{.}(2021)]%
        {han2021adaptive}
\bibfield{author}{\bibinfo{person}{Xueting Han}, \bibinfo{person}{Zhenhuan
  Huang}, \bibinfo{person}{Bang An}, {and} \bibinfo{person}{Jing Bai}.}
  \bibinfo{year}{2021}\natexlab{}.
\newblock \showarticletitle{Adaptive transfer learning on graph neural
  networks}. In \bibinfo{booktitle}{\emph{Proceedings of the 27th ACM SIGKDD
  Conference on Knowledge Discovery \& Data Mining}}.
\newblock


\bibitem[Hu* et~al\mbox{.}(2020)]%
        {Hu2020Strategies}
\bibfield{author}{\bibinfo{person}{Weihua Hu*}, \bibinfo{person}{Bowen Liu*},
  \bibinfo{person}{Joseph Gomes}, \bibinfo{person}{Marinka Zitnik},
  \bibinfo{person}{Percy Liang}, \bibinfo{person}{Vijay Pande}, {and}
  \bibinfo{person}{Jure Leskovec}.} \bibinfo{year}{2020}\natexlab{}.
\newblock \showarticletitle{Strategies for Pre-training Graph Neural Networks}.
  In \bibinfo{booktitle}{\emph{International Conference on Learning
  Representations}}.
\newblock


\bibitem[Hu et~al\mbox{.}(2020b)]%
        {hu2020gpt}
\bibfield{author}{\bibinfo{person}{Ziniu Hu}, \bibinfo{person}{Yuxiao Dong},
  \bibinfo{person}{Kuansan Wang}, \bibinfo{person}{Kai-Wei Chang}, {and}
  \bibinfo{person}{Yizhou Sun}.} \bibinfo{year}{2020}\natexlab{b}.
\newblock \showarticletitle{Gpt-gnn: Generative pre-training of graph neural
  networks}. In \bibinfo{booktitle}{\emph{Proceedings of the 26th ACM SIGKDD
  International Conference on Knowledge Discovery \& Data Mining}}.
\newblock


\bibitem[Hu et~al\mbox{.}(2020a)]%
        {hu2020heterogeneous}
\bibfield{author}{\bibinfo{person}{Ziniu Hu}, \bibinfo{person}{Yuxiao Dong},
  \bibinfo{person}{Kuansan Wang}, {and} \bibinfo{person}{Yizhou Sun}.}
  \bibinfo{year}{2020}\natexlab{a}.
\newblock \showarticletitle{Heterogeneous graph transformer}. In
  \bibinfo{booktitle}{\emph{Proceedings of the web conference 2020}}.
\newblock


\bibitem[Hwang et~al\mbox{.}(2020)]%
        {hwang2020self}
\bibfield{author}{\bibinfo{person}{Dasol Hwang}, \bibinfo{person}{Jinyoung
  Park}, \bibinfo{person}{Sunyoung Kwon}, \bibinfo{person}{KyungMin Kim},
  \bibinfo{person}{Jung-Woo Ha}, {and} \bibinfo{person}{Hyunwoo~J Kim}.}
  \bibinfo{year}{2020}\natexlab{}.
\newblock \showarticletitle{Self-supervised auxiliary learning with meta-paths
  for heterogeneous graphs}. In \bibinfo{booktitle}{\emph{Advances in Neural
  Information Processing Systems}}.
\newblock


\bibitem[Ioannidis et~al\mbox{.}(2022)]%
        {ioannidis2022efficient}
\bibfield{author}{\bibinfo{person}{Vassilis~N Ioannidis},
  \bibinfo{person}{Xiang Song}, \bibinfo{person}{Da Zheng},
  \bibinfo{person}{Houyu Zhang}, \bibinfo{person}{Jun Ma}, \bibinfo{person}{Yi
  Xu}, \bibinfo{person}{Belinda Zeng}, \bibinfo{person}{Trishul Chilimbi},
  {and} \bibinfo{person}{George Karypis}.} \bibinfo{year}{2022}\natexlab{}.
\newblock \showarticletitle{Efficient and effective training of language and
  graph neural network models}.
\newblock \bibinfo{journal}{\emph{arXiv preprint arXiv:2206.10781}}
  (\bibinfo{year}{2022}).
\newblock


\bibitem[Jiang et~al\mbox{.}(2021a)]%
        {jiang2021pre}
\bibfield{author}{\bibinfo{person}{Xunqiang Jiang}, \bibinfo{person}{Tianrui
  Jia}, \bibinfo{person}{Yuan Fang}, \bibinfo{person}{Chuan Shi},
  \bibinfo{person}{Zhe Lin}, {and} \bibinfo{person}{Hui Wang}.}
  \bibinfo{year}{2021}\natexlab{a}.
\newblock \showarticletitle{Pre-training on large-scale heterogeneous graph}.
  In \bibinfo{booktitle}{\emph{Proceedings of the 27th ACM SIGKDD conference on
  knowledge discovery \& data mining}}.
\newblock


\bibitem[Jiang et~al\mbox{.}(2021b)]%
        {jiang2021contrastive}
\bibfield{author}{\bibinfo{person}{Xunqiang Jiang}, \bibinfo{person}{Yuanfu
  Lu}, \bibinfo{person}{Yuan Fang}, {and} \bibinfo{person}{Chuan Shi}.}
  \bibinfo{year}{2021}\natexlab{b}.
\newblock \showarticletitle{Contrastive pre-training of gnns on heterogeneous
  graphs}. In \bibinfo{booktitle}{\emph{Proceedings of the 30th ACM
  International Conference on Information \& Knowledge Management}}.
\newblock


\bibitem[Ke et~al\mbox{.}(2021)]%
        {ke2021jointgt}
\bibfield{author}{\bibinfo{person}{Pei Ke}, \bibinfo{person}{Haozhe Ji},
  \bibinfo{person}{Yu Ran}, \bibinfo{person}{Xin Cui}, \bibinfo{person}{Liwei
  Wang}, \bibinfo{person}{Linfeng Song}, \bibinfo{person}{Xiaoyan Zhu}, {and}
  \bibinfo{person}{Minlie Huang}.} \bibinfo{year}{2021}\natexlab{}.
\newblock \showarticletitle{JointGT: Graph-Text Joint Representation Learning
  for Text Generation from Knowledge Graphs}. In
  \bibinfo{booktitle}{\emph{Findings of the Association for Computational
  Linguistics: ACL-IJCNLP 2021}}.
\newblock


\bibitem[Kipf and Welling(2017)]%
        {kipfsemi}
\bibfield{author}{\bibinfo{person}{Thomas~N Kipf} {and} \bibinfo{person}{Max
  Welling}.} \bibinfo{year}{2017}\natexlab{}.
\newblock \showarticletitle{Semi-Supervised Classification with Graph
  Convolutional Networks}. In \bibinfo{booktitle}{\emph{International
  Conference on Learning Representations}}.
\newblock


\bibitem[Lewis et~al\mbox{.}(2020)]%
        {lewis2020bart}
\bibfield{author}{\bibinfo{person}{Mike Lewis}, \bibinfo{person}{Yinhan Liu},
  \bibinfo{person}{Naman Goyal}, \bibinfo{person}{Marjan Ghazvininejad},
  \bibinfo{person}{Abdelrahman Mohamed}, \bibinfo{person}{Omer Levy},
  \bibinfo{person}{Veselin Stoyanov}, {and} \bibinfo{person}{Luke
  Zettlemoyer}.} \bibinfo{year}{2020}\natexlab{}.
\newblock \showarticletitle{BART: Denoising Sequence-to-Sequence Pre-training
  for Natural Language Generation, Translation, and Comprehension}. In
  \bibinfo{booktitle}{\emph{Proceedings of the 58th Annual Meeting of the
  Association for Computational Linguistics}}.
\newblock


\bibitem[Li et~al\mbox{.}(2021)]%
        {li2021adsgnn}
\bibfield{author}{\bibinfo{person}{Chaozhuo Li}, \bibinfo{person}{Bochen Pang},
  \bibinfo{person}{Yuming Liu}, \bibinfo{person}{Hao Sun},
  \bibinfo{person}{Zheng Liu}, \bibinfo{person}{Xing Xie},
  \bibinfo{person}{Tianqi Yang}, \bibinfo{person}{Yanling Cui},
  \bibinfo{person}{Liangjie Zhang}, {and} \bibinfo{person}{Qi Zhang}.}
  \bibinfo{year}{2021}\natexlab{}.
\newblock \showarticletitle{Adsgnn: Behavior-graph augmented relevance modeling
  in sponsored search}. In \bibinfo{booktitle}{\emph{Proceedings of the 44th
  International ACM SIGIR Conference on Research and Development in Information
  Retrieval}}.
\newblock


\bibitem[Liu et~al\mbox{.}(2019)]%
        {liu2019roberta}
\bibfield{author}{\bibinfo{person}{Yinhan Liu}, \bibinfo{person}{Myle Ott},
  \bibinfo{person}{Naman Goyal}, \bibinfo{person}{Jingfei Du},
  \bibinfo{person}{Mandar Joshi}, \bibinfo{person}{Danqi Chen},
  \bibinfo{person}{Omer Levy}, \bibinfo{person}{Mike Lewis},
  \bibinfo{person}{Luke Zettlemoyer}, {and} \bibinfo{person}{Veselin
  Stoyanov}.} \bibinfo{year}{2019}\natexlab{}.
\newblock \showarticletitle{Roberta: A robustly optimized bert pretraining
  approach}.
\newblock \bibinfo{journal}{\emph{arXiv preprint arXiv:1907.11692}}
  (\bibinfo{year}{2019}).
\newblock


\bibitem[Lu et~al\mbox{.}(2021)]%
        {lu2021learning}
\bibfield{author}{\bibinfo{person}{Yuanfu Lu}, \bibinfo{person}{Xunqiang
  Jiang}, \bibinfo{person}{Yuan Fang}, {and} \bibinfo{person}{Chuan Shi}.}
  \bibinfo{year}{2021}\natexlab{}.
\newblock \showarticletitle{Learning to pre-train graph neural networks}. In
  \bibinfo{booktitle}{\emph{Proceedings of the AAAI conference on artificial
  intelligence}}.
\newblock


\bibitem[Meng et~al\mbox{.}(2022)]%
        {menggnn}
\bibfield{author}{\bibinfo{person}{Yuxian Meng}, \bibinfo{person}{Shi Zong},
  \bibinfo{person}{Xiaoya Li}, \bibinfo{person}{Xiaofei Sun},
  \bibinfo{person}{Tianwei Zhang}, \bibinfo{person}{Fei Wu}, {and}
  \bibinfo{person}{Jiwei Li}.} \bibinfo{year}{2022}\natexlab{}.
\newblock \showarticletitle{GNN-LM: Language Modeling based on Global Contexts
  via GNN}. In \bibinfo{booktitle}{\emph{International Conference on Learning
  Representations}}.
\newblock


\bibitem[Min et~al\mbox{.}(2021)]%
        {min2021recent}
\bibfield{author}{\bibinfo{person}{Bonan Min}, \bibinfo{person}{Hayley Ross},
  \bibinfo{person}{Elior Sulem}, \bibinfo{person}{Amir Pouran~Ben Veyseh},
  \bibinfo{person}{Thien~Huu Nguyen}, \bibinfo{person}{Oscar Sainz},
  \bibinfo{person}{Eneko Agirre}, \bibinfo{person}{Ilana Heinz}, {and}
  \bibinfo{person}{Dan Roth}.} \bibinfo{year}{2021}\natexlab{}.
\newblock \showarticletitle{Recent advances in natural language processing via
  large pre-trained language models: A survey}.
\newblock \bibinfo{journal}{\emph{arXiv preprint arXiv:2111.01243}}
  (\bibinfo{year}{2021}).
\newblock


\bibitem[OpenAI(2023)]%
        {openai2023gpt4}
\bibfield{author}{\bibinfo{person}{OpenAI}.} \bibinfo{year}{2023}\natexlab{}.
\newblock \showarticletitle{GPT-4 Technical Report}.
\newblock \bibinfo{journal}{\emph{arXiv preprint arXiv:2303.08774}}
  (\bibinfo{year}{2023}).
\newblock


\bibitem[Qiu et~al\mbox{.}(2020)]%
        {qiu2020gcc}
\bibfield{author}{\bibinfo{person}{Jiezhong Qiu}, \bibinfo{person}{Qibin Chen},
  \bibinfo{person}{Yuxiao Dong}, \bibinfo{person}{Jing Zhang},
  \bibinfo{person}{Hongxia Yang}, \bibinfo{person}{Ming Ding},
  \bibinfo{person}{Kuansan Wang}, {and} \bibinfo{person}{Jie Tang}.}
  \bibinfo{year}{2020}\natexlab{}.
\newblock \showarticletitle{Gcc: Graph contrastive coding for graph neural
  network pre-training}. In \bibinfo{booktitle}{\emph{Proceedings of the 26th
  ACM SIGKDD international conference on knowledge discovery \& data mining}}.
\newblock


\bibitem[Radford et~al\mbox{.}(2018)]%
        {radford2018improving}
\bibfield{author}{\bibinfo{person}{Alec Radford}, \bibinfo{person}{Karthik
  Narasimhan}, \bibinfo{person}{Tim Salimans}, \bibinfo{person}{Ilya
  Sutskever}, {et~al\mbox{.}}} \bibinfo{year}{2018}\natexlab{}.
\newblock \showarticletitle{Improving language understanding by generative
  pre-training}.
\newblock  (\bibinfo{year}{2018}).
\newblock


\bibitem[Radford et~al\mbox{.}(2019)]%
        {radford2019language}
\bibfield{author}{\bibinfo{person}{Alec Radford}, \bibinfo{person}{Jeffrey Wu},
  \bibinfo{person}{Rewon Child}, \bibinfo{person}{David Luan},
  \bibinfo{person}{Dario Amodei}, \bibinfo{person}{Ilya Sutskever},
  {et~al\mbox{.}}} \bibinfo{year}{2019}\natexlab{}.
\newblock \showarticletitle{Language models are unsupervised multitask
  learners}.
\newblock  (\bibinfo{year}{2019}).
\newblock


\bibitem[Raffel et~al\mbox{.}(2020)]%
        {raffel2020exploring}
\bibfield{author}{\bibinfo{person}{Colin Raffel}, \bibinfo{person}{Noam
  Shazeer}, \bibinfo{person}{Adam Roberts}, \bibinfo{person}{Katherine Lee},
  \bibinfo{person}{Sharan Narang}, \bibinfo{person}{Michael Matena},
  \bibinfo{person}{Yanqi Zhou}, \bibinfo{person}{Wei Li}, {and}
  \bibinfo{person}{Peter~J Liu}.} \bibinfo{year}{2020}\natexlab{}.
\newblock \showarticletitle{Exploring the limits of transfer learning with a
  unified text-to-text transformer}.
\newblock \bibinfo{journal}{\emph{The Journal of Machine Learning Research}}
  (\bibinfo{year}{2020}).
\newblock


\bibitem[Rong et~al\mbox{.}(2020)]%
        {rong2020self}
\bibfield{author}{\bibinfo{person}{Yu Rong}, \bibinfo{person}{Yatao Bian},
  \bibinfo{person}{Tingyang Xu}, \bibinfo{person}{Weiyang Xie},
  \bibinfo{person}{Ying Wei}, \bibinfo{person}{Wenbing Huang}, {and}
  \bibinfo{person}{Junzhou Huang}.} \bibinfo{year}{2020}\natexlab{}.
\newblock \showarticletitle{Self-supervised graph transformer on large-scale
  molecular data}. In \bibinfo{booktitle}{\emph{Advances in Neural Information
  Processing Systems}}.
\newblock


\bibitem[Rosset et~al\mbox{.}(2020)]%
        {rosset2020knowledge}
\bibfield{author}{\bibinfo{person}{Corby Rosset}, \bibinfo{person}{Chenyan
  Xiong}, \bibinfo{person}{Minh Phan}, \bibinfo{person}{Xia Song},
  \bibinfo{person}{Paul Bennett}, {and} \bibinfo{person}{Saurabh Tiwary}.}
  \bibinfo{year}{2020}\natexlab{}.
\newblock \showarticletitle{Knowledge-aware language model pretraining}.
\newblock \bibinfo{journal}{\emph{arXiv preprint arXiv:2007.00655}}
  (\bibinfo{year}{2020}).
\newblock


\bibitem[Schlichtkrull et~al\mbox{.}(2018)]%
        {schlichtkrull2018modeling}
\bibfield{author}{\bibinfo{person}{Michael Schlichtkrull},
  \bibinfo{person}{Thomas~N Kipf}, \bibinfo{person}{Peter Bloem},
  \bibinfo{person}{Rianne Van Den~Berg}, \bibinfo{person}{Ivan Titov}, {and}
  \bibinfo{person}{Max Welling}.} \bibinfo{year}{2018}\natexlab{}.
\newblock \showarticletitle{Modeling relational data with graph convolutional
  networks}. In \bibinfo{booktitle}{\emph{The Semantic Web: 15th International
  Conference, ESWC 2018, Heraklion, Crete, Greece, June 3--7, 2018, Proceedings
  15}}.
\newblock


\bibitem[Shen et~al\mbox{.}(2020)]%
        {shen2020exploiting}
\bibfield{author}{\bibinfo{person}{Tao Shen}, \bibinfo{person}{Yi Mao},
  \bibinfo{person}{Pengcheng He}, \bibinfo{person}{Guodong Long},
  \bibinfo{person}{Adam Trischler}, {and} \bibinfo{person}{Weizhu Chen}.}
  \bibinfo{year}{2020}\natexlab{}.
\newblock \showarticletitle{Exploiting Structured Knowledge in Text via
  Graph-Guided Representation Learning}. In
  \bibinfo{booktitle}{\emph{Proceedings of the 2020 Conference on Empirical
  Methods in Natural Language Processing (EMNLP)}}.
\newblock


\bibitem[Student(1908)]%
        {student1908probable}
\bibfield{author}{\bibinfo{person}{Student}.} \bibinfo{year}{1908}\natexlab{}.
\newblock \showarticletitle{The probable error of a mean}.
\newblock \bibinfo{journal}{\emph{Biometrika}} (\bibinfo{year}{1908}).
\newblock


\bibitem[Sun et~al\mbox{.}(2022)]%
        {sun2022does}
\bibfield{author}{\bibinfo{person}{Ruoxi Sun}, \bibinfo{person}{Hanjun Dai},
  {and} \bibinfo{person}{Adams~Wei Yu}.} \bibinfo{year}{2022}\natexlab{}.
\newblock \showarticletitle{Does {GNN} Pretraining Help Molecular
  Representation?}. In \bibinfo{booktitle}{\emph{Advances in Neural Information
  Processing Systems}}.
\newblock


\bibitem[Wang et~al\mbox{.}(2021)]%
        {wang2021kepler}
\bibfield{author}{\bibinfo{person}{Xiaozhi Wang}, \bibinfo{person}{Tianyu Gao},
  \bibinfo{person}{Zhaocheng Zhu}, \bibinfo{person}{Zhengyan Zhang},
  \bibinfo{person}{Zhiyuan Liu}, \bibinfo{person}{Juanzi Li}, {and}
  \bibinfo{person}{Jian Tang}.} \bibinfo{year}{2021}\natexlab{}.
\newblock \showarticletitle{KEPLER: A Unified Model for Knowledge Embedding and
  Pre-trained Language Representation}.
\newblock \bibinfo{journal}{\emph{Transactions of the Association for
  Computational Linguistics}} (\bibinfo{year}{2021}).
\newblock


\bibitem[Wang et~al\mbox{.}(2019)]%
        {wang2019heterogeneous}
\bibfield{author}{\bibinfo{person}{Xiao Wang}, \bibinfo{person}{Houye Ji},
  \bibinfo{person}{Chuan Shi}, \bibinfo{person}{Bai Wang},
  \bibinfo{person}{Yanfang Ye}, \bibinfo{person}{Peng Cui}, {and}
  \bibinfo{person}{Philip~S Yu}.} \bibinfo{year}{2019}\natexlab{}.
\newblock \showarticletitle{Heterogeneous graph attention network}. In
  \bibinfo{booktitle}{\emph{The world wide web conference}}.
\newblock


\bibitem[Xu et~al\mbox{.}(2019)]%
        {xu2018how}
\bibfield{author}{\bibinfo{person}{Keyulu Xu}, \bibinfo{person}{Weihua Hu},
  \bibinfo{person}{Jure Leskovec}, {and} \bibinfo{person}{Stefanie Jegelka}.}
  \bibinfo{year}{2019}\natexlab{}.
\newblock \showarticletitle{How Powerful are Graph Neural Networks?}. In
  \bibinfo{booktitle}{\emph{International Conference on Learning
  Representations}}.
\newblock


\bibitem[Yang et~al\mbox{.}(2020)]%
        {yang2020heterogeneous}
\bibfield{author}{\bibinfo{person}{Carl Yang}, \bibinfo{person}{Yuxin Xiao},
  \bibinfo{person}{Yu Zhang}, \bibinfo{person}{Yizhou Sun}, {and}
  \bibinfo{person}{Jiawei Han}.} \bibinfo{year}{2020}\natexlab{}.
\newblock \showarticletitle{Heterogeneous network representation learning: A
  unified framework with survey and benchmark}.
\newblock \bibinfo{journal}{\emph{IEEE Transactions on Knowledge and Data
  Engineering}} (\bibinfo{year}{2020}).
\newblock


\bibitem[Yang et~al\mbox{.}(2019)]%
        {yang2019xlnet}
\bibfield{author}{\bibinfo{person}{Zhilin Yang}, \bibinfo{person}{Zihang Dai},
  \bibinfo{person}{Yiming Yang}, \bibinfo{person}{Jaime Carbonell},
  \bibinfo{person}{Russ~R Salakhutdinov}, {and} \bibinfo{person}{Quoc~V Le}.}
  \bibinfo{year}{2019}\natexlab{}.
\newblock \showarticletitle{Xlnet: Generalized autoregressive pretraining for
  language understanding}. In \bibinfo{booktitle}{\emph{Advances in neural
  information processing systems}}.
\newblock


\bibitem[Yao et~al\mbox{.}(2019)]%
        {yao2019graph}
\bibfield{author}{\bibinfo{person}{Liang Yao}, \bibinfo{person}{Chengsheng
  Mao}, {and} \bibinfo{person}{Yuan Luo}.} \bibinfo{year}{2019}\natexlab{}.
\newblock \showarticletitle{Graph convolutional networks for text
  classification}. In \bibinfo{booktitle}{\emph{Proceedings of the Thirty-Third
  AAAI Conference on Artificial Intelligence and Thirty-First Innovative
  Applications of Artificial Intelligence Conference and Ninth AAAI Symposium
  on Educational Advances in Artificial Intelligence}}.
\newblock


\bibitem[Yasunaga et~al\mbox{.}(2022)]%
        {yasunagadeep}
\bibfield{author}{\bibinfo{person}{Michihiro Yasunaga},
  \bibinfo{person}{Antoine Bosselut}, \bibinfo{person}{Hongyu Ren},
  \bibinfo{person}{Xikun Zhang}, \bibinfo{person}{Christopher~D Manning},
  \bibinfo{person}{Percy Liang}, {and} \bibinfo{person}{Jure Leskovec}.}
  \bibinfo{year}{2022}\natexlab{}.
\newblock \showarticletitle{Deep Bidirectional Language-Knowledge Graph
  Pretraining}. In \bibinfo{booktitle}{\emph{Advances in Neural Information
  Processing Systems}}.
\newblock


\bibitem[Yu et~al\mbox{.}(2022)]%
        {yu2022jaket}
\bibfield{author}{\bibinfo{person}{Donghan Yu}, \bibinfo{person}{Chenguang
  Zhu}, \bibinfo{person}{Yiming Yang}, {and} \bibinfo{person}{Michael Zeng}.}
  \bibinfo{year}{2022}\natexlab{}.
\newblock \showarticletitle{Jaket: Joint pre-training of knowledge graph and
  language understanding}. In \bibinfo{booktitle}{\emph{Proceedings of the AAAI
  Conference on Artificial Intelligence}}.
\newblock


\bibitem[Zhu et~al\mbox{.}(2021a)]%
        {zhu2021textgnn}
\bibfield{author}{\bibinfo{person}{Jason Zhu}, \bibinfo{person}{Yanling Cui},
  \bibinfo{person}{Yuming Liu}, \bibinfo{person}{Hao Sun}, \bibinfo{person}{Xue
  Li}, \bibinfo{person}{Markus Pelger}, \bibinfo{person}{Tianqi Yang},
  \bibinfo{person}{Liangjie Zhang}, \bibinfo{person}{Ruofei Zhang}, {and}
  \bibinfo{person}{Huasha Zhao}.} \bibinfo{year}{2021}\natexlab{a}.
\newblock \showarticletitle{Textgnn: Improving text encoder via graph neural
  network in sponsored search}. In \bibinfo{booktitle}{\emph{Proceedings of the
  Web Conference 2021}}.
\newblock


\bibitem[Zhu et~al\mbox{.}(2021b)]%
        {zhu2021transfer}
\bibfield{author}{\bibinfo{person}{Qi Zhu}, \bibinfo{person}{Carl Yang},
  \bibinfo{person}{Yidan Xu}, \bibinfo{person}{Haonan Wang},
  \bibinfo{person}{Chao Zhang}, {and} \bibinfo{person}{Jiawei Han}.}
  \bibinfo{year}{2021}\natexlab{b}.
\newblock \showarticletitle{Transfer learning of graph neural networks with
  ego-graph information maximization}. In \bibinfo{booktitle}{\emph{Advances in
  Neural Information Processing Systems}}.
\newblock


\end{thebibliography}

\newpage
\appendix
\onecolumn

\section{Data Statistics}
\label{app:datastats}

\subsection{Data Statistics of \amzcore}
\label{app:datastats_amzcore}

\begin{table}[htbp]
\centering
\caption{Statistics of internal datasets of \amzcore (including sub-sampled a large graph corpus for pre-training and three application graphs).}
    \resizebox{0.95\linewidth}{!}{
    \begin{threeparttable}
    \begin{tabular}{ l | l  l | l }
    \toprule
     \multirow{2}{*}{\amzcore} & \multicolumn{2}{c|}{\# Nodes} &  \multirow{2}{*}{\# Edges (query $\rightarrow$ product)}  \\
     & product & query &  \\
     \midrule
     \midrule
     The large graph corpus (sub-sampled) & $1,719,361$ & $11,165,437$ & add: $56,087,935$, click: $147,985,529$, consume: $252,818$,  purchase: $25,861,965$ \\
     \midrule
    \searchctr & $65,751$ & $1,852,895$ & ads-click: $6,385,003$ \\
    \esci & $1,660,560$ &  $209,874$ & match: $2,548,873$ \\
    \qpt  &  $1,121,831$ &  $552,789$ & click: $10,379,242$ \\
    \bottomrule
    \end{tabular}
    \end{threeparttable}
    }
\label{tab:datastats_amzcore}
\end{table}

\subsection{Data Statistics of \amzreviews}
\label{app:datastats_amzreviews}

\begin{table*}[htbp]
\centering
\caption{Statistics of public datasets of \amzreviews (including a large graph corpus and two applications).}
    \resizebox{0.8\linewidth}{!}{
    \begin{tabular}{ l | l l | l }
    \toprule
     \multirow{2}{*}{\amzreviews} & \multicolumn{2}{c|}{\# Nodes} &  \multirow{2}{*}{\# Edges} \\
     & asin & reviewText \\
     \midrule
     \midrule
     The large graph & $2,781,996$ & $2,590,064$ & asin--coview--asin: $17,491,876$, reviewText--review--asin: $2,805,401$, \\
     corpus & & & reviewText--cowrite--reviewText: $35,769,994$ \\
     \midrule
    \copurchase & $424,716$ & $13,032$ & asin--cobuy--asin: $1,161,159$, reviewText--review--asin: $278,063$ \\
    \acat &  $694,524$ & -- & asin--cobuy--asin: $1,738,293$ \\
    \bottomrule
    \end{tabular}
    }
\label{tab:datastats_amareviews}
\end{table*}

\section{More Experimental Details}
\label{app:moredetails}

\subsection{Elapsed Time of Pre-training Strategies}
\label{app:pretraining_time}
Due to the internal legal policy, we could only show relative throughput performances between the two pre-training methods.
\paragraph{\textbf{Graph-aware LM Pre-training}}
The time elapsed for graph-aware LM pre-training is 100\%.
\paragraph{\textbf{GNN-based Graph-aware LM Pre-training}}
The time elapsed for GNN-based graph-aware LM pre-training is 977\% (100\% for graph-aware LM pre-training, 8\% for GNN warming up, and 869\% for co-training LMs with a GNN aggregator). Even if the graph-aware LM pre-training is not trained till 100\%, the total elapsed time of \galmco is more than 877\%.

\subsection{Hyper-parameter Setting}
\label{app:hyperparas}

\begin{table}[htbp]
\centering
\caption{Hyper-parameters for models and experiments.}
    \resizebox{\linewidth}{!}{
    \begin{tabular}{l l |  c c  c  c | c  c c}
    \toprule
    \multirow{3}{*}{Category} & \multirow{3}{*}{Hyperparameter} & \multicolumn{4}{c|}{\amzcore} & \multicolumn{3}{c}{\amzreviews} \\
    & & The large graph & \multirow{2}{*}{\searchctr} & \multirow{2}{*}{\esci} & \multirow{2}{*}{\qpt} & The large graph & \multirow{2}{*}{\copurchase} & \multirow{2}{*}{\acat} \\
    & & corpus & & & & corpus \\
     \midrule
     \midrule
    \multirow{4}{*}{Model architecture} & Number of GNN aggregator layers & \centering{1-2} & 1 & 1 & 1 & \centering{1-2} & 1 & 1 \\
        & Dimension of GNN aggregator hidden layers & \centering{256} & 256 & 256 & 256 & \centering{256} & 256 & 256 \\
        & Number of attention heads of RGAT & \centering{4} & 4 & 4 & 4 & \centering{4} & 4 & 4 \\
        & Type of decoders & \centering{1-layer MLP} & 1-layer MLP & 1-layer MLP & 1-layer MLP & \centering{1-layer MLP} & 1-layer MLP & 1-layer MLP \\
    \midrule
    \multirow{6}{*}{Experimental setups} & Learning rate of parameters of LMs & 1e-8 & 1e-7 & 1e-6 & 1e-7 & 1e-8 & 1e-7 & 1e-7 \\
        & Learning rate of parameters of GNNs & 5e-4 & 5e-4 & 1e-4 & 5e-4 & 5e-4 & 5e-4 & 5e-4 \\
        & Optimizer &  Adam & Adam & Adam & Adam & Adam & Adam & Adam \\
        & Batch size for training  &  512  &  512  &  512  &  512  &  512  &  512  &  512  \\
        & Batch size for evaluation   &  1024  &  1024  &  1024  &  1024  &  1024  &  1024  &  1024  \\
        & Max number of tokens & \centering{256} & 256 & 256 & 256 & \centering{256} & 256 & 256 \\
    \bottomrule
    \end{tabular}
    }
\label{tab:hyperparas}
\end{table}

\newpage
\subsection{Original Results for Significant Analysis}
\label{app:sigresults}

\begin{table*}[htbp]
\centering
\caption{The original results (micro-F1 and accuracy) of five repetitions on \acat for significant analysis.}
\resizebox{0.83\textwidth}{!}{
    \begin{tabular}{ l | c c c c c | c c c c c }
    \toprule
    \multirow{2}{*}{Models} & \multicolumn{10}{c}{\acat} \\
     & \multicolumn{5}{c}{\footnotesize(macro-F1)} & \multicolumn{5}{c}{\footnotesize(accuracy)} \\
    \midrule
    \midrule
    \hf+\rgat & 0.7038 &  0.7044 & 0.6977 & 0.7051 & 0.7014
            & 0.8133 & 0.8127  & 0.8149 &  0.8101 & 0.8108 \\
    \hfft+\rgat & 0.7479 & 0.7486 & 0.7455 & 0.7455 & 0.7458 
            & 0.8435 & 0.8424  & 0.8406 & 0.8416 & 0.8415 \\
    \midrule
    \galmrgat & 0.7317 & 0.7314 & 0.7319 & 0.7287 & 0.7300 
            & 0.8311 & 0.8330 & 0.8321  & 0.8311 & 0.8302\\
    \galmrgatft & 0.7491 & 0.7490 & 0.7449 & 0.7473 & 0.7466 
            & 0.8427 & 0.8437 & 0.8436 & 0.8427 & 0.8437 \\
    \galmrgatftplus & 0.7813 & 0.7838 & 0.7838 & 0.7831 & 0.7856
            & 0.8676 & 0.8660 & 0.8671 & 0.8672 & 0.8670 \\
    \bottomrule
    \end{tabular}}
\label{tab:siganalysis}
\end{table*}

\end{document}